\documentclass{article}

\usepackage{arxiv}

\usepackage[utf8]{inputenc} 
\usepackage[T1]{fontenc}    
\usepackage{hyperref}       
\usepackage{url}            
\usepackage{booktabs}       
\usepackage{amsfonts}       
\usepackage{nicefrac}       
\usepackage{microtype}      
\usepackage{lipsum}

\usepackage{amsmath}
\usepackage{amssymb}
\usepackage{amsfonts}
\usepackage{amstext}

\usepackage[utf8]{inputenc}
\usepackage{chngcntr}
\usepackage{appendix}
\usepackage{amsthm}

\usepackage{xcolor, etoolbox}

\usepackage{hyperref}

\hypersetup{
   colorlinks=true,
   urlbordercolor={1 1 1},
   citecolor={blue}
}

\usepackage[T1]{fontenc}    
\usepackage{hyperref}       
\usepackage[left]{lineno}

\usepackage{url}            
\makeatletter
\g@addto@macro{\UrlBreaks}{\UrlOrds}
\makeatother

\usepackage{booktabs}       
\usepackage{amsfonts}       
\usepackage{nicefrac}       
\usepackage{microtype}      
\usepackage{lipsum}

\usepackage{amsfonts}
\usepackage{relsize}
\usepackage{comment}
\usepackage{listings}

\usepackage{multirow}
\usepackage{tabulary,capt-of}
\usepackage{adjustbox}
\usepackage{diagbox}

\usepackage{algpseudocode}
\usepackage{multicol}
\usepackage{caption}

\usepackage{tabularx}
\usepackage{pdfpages}
\usepackage{graphicx}
\usepackage{caption, subcaption}
\usepackage{booktabs}

\usepackage{mathtools}
\newtagform{blue}{\color{red}(}{)}
\newtagform{redandblue}[\textcolor{red}]{\color{red}(}{)}

\usetagform{blue}

\newcounter{unnumber}

\usepackage{multirow}
\usepackage{colortbl}
\usepackage{multicol}

\usepackage{mathpazo}
\usepackage{physics}

\usepackage[ruled,vlined]{algorithm2e}

\title{Some notes concerning a generalized KMM-type optimization method for density ratio estimation}

\author{
 Cristian Daniel Alecsa \\
  Romanian Institute of Science and Technology \\
  Technical University of Cluj-Napoca\\
  Cluj-Napoca, Romania \\
  \texttt{alecsa.cd@gmail.com} \\ \\
  \textbf{ORCID }: $0000$-$0003$-$2324$-$7711$
}

\begin{document}

\maketitle
\begin{abstract}
In the present paper we introduce new optimization algorithms for the task of density ratio estimation. More precisely, we consider extending the well-known KMM method using the construction of a suitable loss function, in order to encompass more general situations involving the estimation of density ratio with respect to subsets of the training data and test data, respectively. The associated codes can be found at \url{https://github.com/CDAlecsa/Generalized-KMM}.
\end{abstract}

\section{Introduction}\label{sec:G-KMM_Introduction}
In statistical data processing, the comparison of two distributions is of paramount importance. In general, the problem of assessing whether two probability distributions are equivalent or not is addressed through the so-called two-sample tests. There exists classical methods that tackle this issue, such as the $\textit{t-test}$ which compares the means of two Gaussian distributions with common variance, and the well-known non-parametric $\textit{Kolmogorov-Smirnov test}$. Recently, in \cite{Gretton_MMD}, Gretton et. al. introduced the \emph{maximum mean discrepancy} (MMD) statistic which compares the similarities through a positive-defined kernel across two samples in an universal reproducing kernel Hilbert space (universal RKHS), and is commonly used for multivariate two-sample testing. In \cite{DeepMMD}, Kirchler et. al. extended the MMD statistic by using a neural network trained on an auxiliary dataset which defines the so-called \emph{deep maximum mean discrepancy} (DMMD) statistic, where the mapping from the input domain to the network's last hidden layer is utilized as the kernel used in the MMD statistic. \\
A different approach to the problem of two-sample testing is based upon the evaluation of a divergence between two distributions, such as the $\textit{f-divergence}$ which includes the case of the $\textit{Kullback-Leibler divergence}$ and the $\textit{Pearson divergence}$, respectively. Due to the fact that the density estimation is a hard task, a practical approach to the divergence estimation is to directly approximate the density ratio function. One of the most well known method is the \emph{kernel mean matching} (KMM) algorithm from \cite{Gretton_KMM} based on infinite-order moment matching, and which represents the MMD statistic in the particular case when a distribution is weighted accordingly to the density ratio model. There are several extensions of the KMM method such as the \emph{Ensemble KMM} introduced in \cite{EnsembleKMM} where, instead of a single train-test split, one uses multiple non-overlapping test datasets and a single train dataset. A more generalized version was introduced in \cite{EfficientSamplingKMM} where the main idea is to divide the training data due to the fact that the \emph{Ensemble KMM} is not suitable with large training datasets. Consequently, this method which we will call it \emph{Efficient Sampling KMM}, takes a bootstrap approach for the training data and then merges the results with an aggregation process. A combination between the aforementioned two methodologies was done in \cite{Haque} where Haque and his coauthors considered a KMM-type density ratio estimation called SKMM based on using bootstrap generation method for the training data and a partitioning of the test data. An extension of the classical \emph{KMM} method is the neural network technique introduced \cite{NeuralNetwork_KMM} where the loss function is the actual MMD objective and the bandwidth of the underlying kernel is considered as a hyper-parameter. Due to the inherent flexibility of neural architectures and that the training is done on randomized batches this Deep Learning approach outperforms the classical KMM algorithms. \\
Different alternatives to MMD-based methods mostly rely on solving different \emph{optimization problems}. Such examples are the \emph{unconstrained least squares importance fitting} (uLSIF) from \cite{uLSIF} and the \emph{relative uLSIF} (RuLSIF) method introduced in \cite{RuLSIF}, where in the latter one the density ratio model involves a mixture of densities. The uLSIF method was successfully employed in \cite{LeastSquaresTwoSampleTest} for the comparison of  distributions using a permutation test approach. In more detail, by utilizing a weighted Gaussian kernel model at test samples, the weights of the density ratio model are learned through a \emph{quadratic optimization problem}, after which the \emph{Pearson divergence} is employed in order to compare the train and test distributions. A general method encompassing the constrained variant of the uLSIF method, namely LSIF, is the Bregman formulation given in \cite{Bregman_DRE}. The unified method which utilizes the Bregman divergence contains as a particular case the well-known \emph{KL importance estimation procedure} (KLIEP) from \cite{KLIEP} for which the \emph{objective function} is given with respect to the test samples, while the constraints depend on the training samples. When the true density ratio is approximated through a linear or kernel model, then one obtains a \emph{convex optimization problem with constraints}. On the other hand, for the situation when the choice of the density ratio model is the \emph{log-linear model} as in \cite{LogKLIEP} then the underlying density fitting framework reduces to a \emph{unconstrained convex optimization problem} and therefore the global solution can be obtained by iterative methods such as \emph{gradient descent}. \\
As previously mentioned, the KMM method introduced in \cite{NeuralNetwork_KMM} uses the MMD statistic as the \emph{loss function} for the density ratio approximation which is modeled through a neural network. A similar approach was developed recently in \cite{NeuralNetwork_ChangePointDetection} in which a neural-type approach was developed for the RuLSIF method in the setting of change point detection. In both these works, the true density ratio is represented as a neural network and the learning of such a network depends on a loss function suitable for density ratio estimation. A different approach for the RuLSIF method is the one from \cite{MetaLearning_DRE} where the density ratio is not directly represented as a neural network but is considered as a weighted feed-forward neural model (instead of a weighted kernel model). By employing the RuLSIF approach one finds at each iteration the weights as the global optimum solution for the quadratic optimization problem. After finding the weights, the classical backpropagation algorithm is applied to the density ratio model with respect to the parameters of the feed-forward neural model. In addition to this, the aim of the method introduced in \cite{MetaLearning_DRE} is to estimate the density ratio from a few training samples, by using instances in different but related datasets (also called source datasets). \\
As we formerly emphasized, the framework regarding the density ratio models \emph{Ensemble KMM} from \cite{EnsembleKMM} and \emph{Efficient Sampling KMM} from \cite{EfficientSamplingKMM} depends on multiple train or test datasets. On the other hand, the density fitting methodology proposed in \cite{MultiDistribution_DRE} (which we shall briefly call it \emph{MultiDistribution DRE}) is related to the idea that one has access to i.i.d. samples from multiple distributions. As a particular case, this can be perceived as having multiple test datasets and a single reference train dataset. The purpose is to estimate the density ratios between all pairs of distributions. This can be efficiently done by employing the Bregman divergence with respect to the reference density function and thus optimizing a vector density ratio. Consequently, the optimization function can be written as a sum of multiple objective mappings, where each of them depends on a particular density ratio component. Accordingly, this approach generalizes the LSIF and KLIEP optimization algorithms to the so-called Multi-LSIF and Multi-KLIEP methods, respectively. \\
There are various alternatives to the aforementioned density ratio methods. The most well known approach to the previously mentioned techniques is the probabilistic density fitting method where, as described in \cite{DRE_BOOK}, one learns a probabilistic classifier that separates the train and test samples. The methodology behind these classification algorithms is that the density ratio is approximated by the ratio of the sample sizes multiplied by the class posterior probabilities, the latter ones being obtained from the classifier's output. Furthermore, the main advantage of the probabilistic classification technique is that it is easy to implement it in a real world situation.

\subsection{Structure of the paper}
The aim of section \eqref{sec:G-KMM_Motivation} is to present in a step-by-step manner the main stimulus behind our \texttt{Generalized KMM} method. For this, we will investigate an approach for devising a \emph{quadratic optimization problem with constraints} based on the situation of multiple non-overlapping training datasets, along with the case of multiple non-overlapping test datasets. Our algorithmic framework is different than the methodologies from \emph{Ensemble KMM} introduced in \cite{EnsembleKMM} and \emph{Efficient Sampling KMM} from \cite{EfficientSamplingKMM} since our technique is constructed using a loss-function approach. In section \eqref{sec:G-KMM_ProposedOptMethod} we will actually construct our optimizer. By introducing an extended density ratio function using mixtures of probability densities, our technique is based upon the construction of a non-negative loss mapping which attains its minimum value of $0$ under the true density ratio. The empirical version is obtained when one uses an approximate linear kernel model using the points of the whole test dataset, by utilizing some constraints suitable to numerical implementations. In section \eqref{sec:G-KMM_Experiments} we present some numerical simulations developed with a custom implementation made in \texttt{Python} regarding the comparison of probability densities under the learned density ratio weights, along with importance reweighting examples. Finally, in the last section \eqref{sec:G-KMM_Conclusions}, we discuss about the advantages of our generalized method along with the underlying limitations.

\section{Motivation}\label{sec:G-KMM_Motivation}
Let's consider two sample sets $\mathcal{X} = \lbrace x_i \, | \, x_i \in \mathbb{R}^d \rbrace_{i = 1}^{n}$ and $\mathcal{X}^\prime = \lbrace x^{\prime}_{j} \, | \, x^{\prime}_{j} \in \mathbb{R}^d \rbrace_{j = 1}^{n^\prime}$ such that $\mathcal{X} \overset{i.i.d.}{\sim} P$ and $\mathcal{X}^\prime \overset{i.i.d.}{\sim} P^\prime$, where $P$ and $P^\prime$ are probability distributions with densities $p, p^\prime$, respectively. \\
The MMD (maximum mean discrepancy) statistic between $\mathcal{X}$ and $\mathcal{X}^\prime$ is defined as 
\begin{align*}
MMD_{\phi, \psi}(\mathcal{X}, \mathcal{X}^\prime) = \| \mathbb{E}_{p^\prime(x)} [\psi(x)] - \mathbb{E}_{p(x)}[\phi(x)] \|^2,    
\end{align*}
where $\phi, \psi: \mathbb{R}^d \to \mathbb{R}^p$ are two given feature maps. By defining the density ratio function $r(x) = \dfrac{p(x)}{p^\prime(x)}$, where we assume that $p^\prime(x) > 0$ for all $x$, and choosing $\psi(x) = r(x) \phi(x)$, then we obtain the loss function used in the KMM (kernel mean matching) approach from \cite{Gretton_KMM} with respect to an approximation $\hat{r}$ of $r$:
\begin{align*}
MMD_{\phi, \widehat{r} \phi}(\mathcal{X}, \mathcal{X}^\prime) = \| \mathbb{E}_{p^\prime(x)} [\hat{r}(x)\phi(x)] - \mathbb{E}_{p(x)}[\phi(x)] \|^2
\end{align*}
By using that 
$$ 1 = \int p(x) dx = \int r(x) p^\prime(x) dx = \mathbb{E}_{p^\prime(x)}[r(x)] $$
and taking into account the above objective function, one obtains the following \emph{optimization problem with constraints} (that needs to be solved by considering an approximation of the density ratio model $\hat{r}$):
\begin{align}\label{KMM}\tag{OptPb-KMM}
\begin{cases}
&\min\limits_{\widehat{r}} \, \| \mathbb{E}_{p^\prime(x)} [\hat{r}(x)\phi(x)] - \mathbb{E}_{p(x)}[\phi(x)] \|^2 \\ \\
&\text{subject to}
\begin{cases}
\hat{r}(x) \geq 0 \text{ for all } x \nonumber\\
\mathbb{E}_{p^\prime(x)}[\hat{r}(x)] = 1
\end{cases}
\end{cases}
\end{align}
For simplifying the formulation of \eqref{KMM}, we introduce the kernel map $K: \mathbb{R}^d \times \mathbb{R}^d \to \mathbb{R}$ such that $K(x, y) = \langle \phi(x), \phi(y) \rangle$. Furthermore, let's consider $h \in \mathbb{R}^{n^\prime}$ such that $h_j = \dfrac{n^\prime}{n} \sum\limits_{i = 1}^{n} K(x^\prime_j, x_i)$ for $j = \lbrace 1, \ldots, n^\prime \rbrace$. At the same time, define $H \in \mathbb{R}^{n^\prime \times n^\prime}$ such that $H_{jk} = K(x^\prime_j, x^\prime_k)$ for $j, k \in \lbrace 1, \ldots, n^\prime \rbrace$, along with $r_{\mathcal{X}^\prime} \in \mathbb{R}^{n^\prime \times 1}$, where $r_{\mathcal{X}^\prime} = (r(x^\prime_1), \ldots, r(x^\prime_{n^\prime}))^T$. If we define $\hat{r}(x)$ as a model approximating the true density ratio $r(x)$, and ignoring irrelevant constants with respect to $\hat{r}(x)$ then \eqref{KMM} becomes the following \emph{quadratic optimization problem with constraints}:
\begin{align*}
\begin{cases}
&\min\limits_{\widehat{r}_{\mathcal{X}^\prime}} \, \left( \dfrac{1}{2} \hat{r}_{\mathcal{X}^\prime}^T H \hat{r}_{\mathcal{X}^\prime} - h^T \hat{r}_{\mathcal{X}^\prime} \right) \\
&\text{subject to}
\begin{cases}
(\hat{r}_{\mathcal{X}^\prime})_j \geq 0 \text{ for } j \in \lbrace 1, \ldots, n^\prime \rbrace \\ \\
\dfrac{1}{n^\prime} \sum\limits_{j = 1}^{n^\prime} (\hat{r}_{\mathcal{X}^\prime})_j = 1
\end{cases}
\end{cases}
\end{align*}
A different kind of density ratio technique is the RuLSIF method from \cite{RuLSIF} which uses a generalization of the density ratio i.e., the $\alpha$-relative density ratio $r_\alpha(x) = \dfrac{p(x)}{q_\alpha(x)}$ for $\alpha \in [0, 1)$, where $q_\alpha(x)$ represents the $\alpha$-mixture density of $p(x)$ and $p^\prime(x)$ i.e., $q_\alpha(x) = \alpha p(x) + (1 - \alpha) p^\prime(x)$. In the RuLSIF method, one models the true density ratio as a linear kernel method with weights $w \in \mathbb{R}^{n \times 1}$, namely $\hat{r}(x) = w^T k(x)$, where $k(x) = \left(K(x, x_1), \ldots, K(x, x_n) \right)^T \in \mathbb{R}^{n \times 1}$ and $K(x, y) = \exp \left( - \dfrac{\| x - y \|^2}{2 \tilde{\sigma}^2} \right)$ is the corresponding Gaussian kernel with the variance $\tilde{\sigma}^2$, respectively. Let's define $H \in \mathbb{R}^{n \times n}$ and $h \in \mathbb{R}^{n \times 1}$, namely $H_{l, k} = \dfrac{\alpha}{n} \sum\limits_{i = 1}^{n} K(x_i, x_l) K(x_i, x_k) + \dfrac{1 - \alpha}{n^\prime} \sum\limits_{j = 1}^{n^\prime} K(x^\prime_j, x_l) K(x^\prime_j, x_k) $ and $h_l = \dfrac{1}{n} \sum\limits_{i = 1}^{n} K(x_i, x_l)$, where $l, k \in \lbrace 1, \ldots, n \rbrace$. By ignoring constants irrelevant to the weights $w$ and using a regularization parameter $\lambda$, we obtain the following \emph{unconstrained regularized quadratic optimization problem} corresponding to the RuLSIF loss:
\begin{align}\label{RuLSIF}\tag{OptPb-RuLSIF}
\min_w \left( \dfrac{1}{2} w^T H w - h^T w + \dfrac{\lambda}{2} w^T w \right)
\end{align}
In order to generalize the KMM approach to multiple sample sets, we observe that the objective function belonging to \eqref{KMM} leads to
\begin{align}\label{eq:1}
\mathbb{E}_{p(x)}[\phi(x)] = \int \phi(x) p(x) dx = \int r(x) \phi(x) p^\prime(x) dx = \mathbb{E}_{p^\prime(x)}[r(x)\phi(x)],  
\end{align}
therefore, under the true density ratio model $r(x)$, the loss function is minimized. We will use this simple technique for extending in a precise manner the KMM algorithm to multiple sample sets. \\
\vskip+0.25cm
$\bullet \,$ The first case we investigate is when we have, for $i \in \lbrace 1, \ldots, N \rbrace$, the sets $\mathcal{X}_i = \lbrace x_{k,(i)} \, | \, x_{k, (i)} \in \mathbb{R}^d \rbrace_{k = 1}^{n_i}$ and $\mathcal{X}^\prime = \lbrace x^{\prime}_{j} \, | \, x^{\prime}_{j} \in \mathbb{R}^d \rbrace_{j = 1}^{n^\prime}$ such that $\mathcal{X}_i \overset{i.i.d.}{\sim} P_{i}$ and $\mathcal{X}^\prime \overset{i.i.d.}{\sim} P^\prime$, where $P_{i}$ and $P^\prime$ are probability distributions with densities $p_{i}, p^\prime$, respectively. If we consider $\mathcal{X}^\prime$ to be the training dataset and $\mathcal{X}_i$ to represent the non-overlapping partitions of a given test dataset $\mathcal{X}$, then the technique proposed in \emph{Ensemble KMM} uses the fact that $p(x \in \mathcal{X}) = \sum\limits_{i = 1}^{N} \dfrac{n_i}{n} p(x \, | \, x \in \mathcal{X}_i)$, where $n = \sum\limits_{j = 1}^{N} n_j$. Therefore, the probability associated to the test sample set $\mathcal{X}$ can be written as a mixture between non-overlapping partitions with the weights given by the ratio between the size of the partition and the total size of the test dataset. This is in accordance with the definition from the formulation of \emph{Ensemble KMM} of the density ratio corresponding to $\mathcal{X}$ and $\mathcal{X}^\prime$ which is given by a mixture of density ratios between $\mathcal{X}_i$ and $\mathcal{X}^\prime$, respectively. By dropping the notation of conditional probability density concerning the partitions of $\mathcal{X}$, let's consider the general case when $r(x) = \sum\limits_{i = 1}^N \omega_i r_i(x)$ where $r_i(x) = \dfrac{p_i(x)}{p^\prime(x)}$ for $i \in \lbrace 1, \ldots, N \rbrace$ and where the weights satisfy $\sum\limits_{i = 1}^N \omega_i = 1$ with $\omega_i \in [0, 1]$ for $i \in \lbrace 1, \ldots, N \rbrace$. Therefore
\begin{align*}
r(x) = \sum\limits_{i = 1}^N \omega_i r_i(x) = \sum\limits_{i = 1}^N \omega_i \dfrac{p_i(x)}{p^\prime(x)} = \dfrac{\sum\limits_{i = 1}^N \omega_i p_i(x)}{p^\prime(x)} = \dfrac{p(x)}{p^\prime(x)},  
\end{align*}
where $p(x)$ represents the mixture density defined as $p(x) = \sum\limits_{i = 1}^N \omega_i p_i(x)$. Inspired by the identity \eqref{eq:1} from the case of KMM, we infer the loss function which is minimized under the true density ratio $r(x)$:
\begin{align*}
\mathbb{E}_{p^\prime(x)}[r(x)\phi(x)] &= \int r(x) \phi(x) p^\prime(x) dx = \int \left( \sum\limits_{i = 1}^{N} \omega_i p_i(x) \right) \phi(x) dx \\
&= \sum\limits_{i = 1}^{N} \omega_i \int p_i(x) \phi(x) dx = \sum\limits_{i = 1}^{N} \omega_i \mathbb{E}_{p_i(x)}[\phi(x)],    
\end{align*}
thus we consider the following loss function which needs to be solved with respect to the approximate model $\hat{r}$ of the density ratio $r$:
\begin{align}\label{eq:2}
\mathcal{L} = \left\| \mathbb{E}_{p^\prime(x)}[\hat{r}(x)\phi(x)] - \sum\limits_{i = 1}^{N} \omega_i \mathbb{E}_{p_i(x)}[\phi(x)] \right\|^2
\end{align}
\vskip+0.25cm
An alternative of the above computations is to consider the approach of \emph{MultiDistribution DRE} where we define as before, for $i \in \lbrace 1, \ldots, N \rbrace$, $r_i(x) = \dfrac{p_i(x)}{p^\prime(x)}$. In this case we compute for every $i \in \lbrace 1, \ldots, N \rbrace$ the following:
\begin{align*}
\mathbb{E}_{p^\prime(x)}[r_i(x)\phi(x)] = \int r_i(x) \phi(x) p^\prime(x) dx = \int \phi(x) p_i(x) dx = \mathbb{E}_{p_i(x)}[\phi(x)],    
\end{align*}
hence we can define the $i^{th}$ loss mapping with respect to the approximation $\hat{r}_i$ of $r_i$:
\begin{align*}
\mathcal{L}_i = \| \mathbb{E}_{p^\prime(x)}[\hat{r}_i(x)\phi(x)] - \mathbb{E}_{p_i(x)}[\phi(x)] \|^2,    
\end{align*}
therefore one can propose the mixture loss function $\mathcal{L}$ where the weights $w_i$ represent the contribution of each particular loss function $\mathcal{L}_i$, namely
\begin{align}\label{eq:3}
\mathcal{L} = \sum\limits_{i = 1}^{N} \omega_i \mathcal{L}_i = \sum\limits_{i = 1}^{N} \omega_i \| \mathbb{E}_{p^\prime(x)}[\hat{r}_i(x)\phi(x)] - \mathbb{E}_{p_i(x)}[\phi(x)] \|^2.    
\end{align}
It is worth pointing out that the loss function \eqref{eq:3} resembles the approach of \emph{Ensemble KMM} structure, where one solves simultaneously $N$ optimization problems, with the condition that the objective function of the $i^{th}$ problem is related to the approximation $\hat{r}_i(x)$ of its associated density ratio model $r_i$. \\
\vskip+0.25cm
$\bullet \,$ Now we turn our attention to our second case which we shall analyze it, where from a practical point of view one has multiple non-overlapping training datasets and a single test dataset. In order to do this we consider, for $i \in \lbrace 1, \ldots, N^\prime \rbrace$, the sets $\mathcal{X} = \lbrace x_k \, | \, x_k \in \mathbb{R}^d \rbrace_{k = 1}^{n}$ and $\mathcal{X}^\prime_i = \lbrace x^{\prime}_{j, (i)} \, | \, x^{\prime}_{j, (i)} \in \mathbb{R}^d \rbrace_{j = 1}^{n^\prime_i}$ such that $\mathcal{X} \overset{i.i.d.}{\sim} P$ and $\mathcal{X}^\prime_{i} \overset{i.i.d.}{\sim} P^\prime_{i}$, where $P$ and $P^\prime_{i}$ are probability distributions with densities $p, p^\prime_{i}$, respectively. In a similar fashion with the previous case, let's consider $r_i(x) = \dfrac{p(x)}{p_i^\prime(x)}$ and the weights $\Tilde{\omega}_i \in [0, 1]$, for each $i \in \lbrace 1, \ldots, N^\prime \rbrace$ such that $\sum\limits_{i = 1}^{N^\prime} \Tilde{\omega}_i = 1$. Then, it follows that
\begin{align*}
\mathbb{E}_{p(x)}[\phi(x)] &= \int \phi(x) p(x) dx = \int \phi(x) p(x) \left( \sum\limits_{i = 1}^{N^\prime} \Tilde{\omega}_i \right) dx \\
&= \int \phi(x) \left( \Tilde{\omega}_1 p(x) + \ldots + \Tilde{\omega}_{N^\prime} p(x) \right) dx \\
&= \int \phi(x) \left( \Tilde{\omega}_1 r_1(x) p^\prime_1(x) + \ldots + \Tilde{\omega}_{N^\prime} r_{N^\prime}(x) p^\prime_{N^\prime}(x) \right) dx \\
&= \int \phi(x) \left( \sum\limits_{i=1}^{N^\prime} \Tilde{\omega}_i r_i(x) p_i^\prime (x) \right) dx = \sum\limits_{i=1}^{N^\prime} \Tilde{\omega}_i \int \phi(x) r_i(x) p_i^\prime(x) dx \\
&= \sum\limits_{i=1}^{N^\prime} \Tilde{\omega}_i \mathbb{E}_{p_i^\prime(x)}[r_i(x)\phi(x)],
\end{align*}
hence it is natural to propose the following loss function:
\begin{align}\label{eq:4}
\mathcal{L} = \left\| \sum\limits_{i=1}^{N^\prime} \Tilde{\omega}_i \mathbb{E}_{p_i^\prime(x)}[\hat{r}_i(x)\phi(x)] - \mathbb{E}_{p(x)}[\phi(x)] \right\|^2
\end{align}
\vskip+0.25cm
For the previous case we shall present an alternative method in order to infer a suitable loss function. We proceed by considering, for each $i \in \lbrace 1, \ldots, N^\prime \rbrace$ the weights $\omega_i \in [0, 1]$ that satisfy $\sum\limits_{i = 1}^{N^\prime} \omega_i = 1$. Along with these we define the mixture probability density $p^\prime(x) = \sum\limits_{i=1}^{N^\prime} \omega_i p_i^\prime(x)$ and the corresponding true density ratio $r(x) = \dfrac{p(x)}{p^\prime(x)}$. Then, it follows that
\begin{align*}
\mathbb{E}_{p(x)}[\phi(x)] &= \int \phi(x) p(x) dx = \int \phi(x) r(x) p^\prime(x) dx = \int \phi(x) r(x) \left( \sum\limits_{i=1}^{N^\prime} \omega_i p_i^\prime(x) \right) dx \\
&= \sum\limits_{i=1}^{N^\prime} \omega_i \int \phi(x) r(x) p_i^\prime(x) dx = \sum\limits_{i=1}^{N^\prime} \omega_i \mathbb{E}_{p_i^\prime(x)} [r(x) \phi(x)],
\end{align*}
which defines the following loss function:
\begin{align}\label{eq:5}
\mathcal{L} = \left\| \sum\limits_{i=1}^{N^\prime} \omega_i \mathbb{E}_{p_i^\prime(x)} [\hat{r}(x) \phi(x)] - \mathbb{E}_{p(x)}[\phi(x)] \right\|^2
\end{align}
\vskip+0.25cm
In what follows we will show an equivalence between \eqref{eq:4} and \eqref{eq:5} under certain assumptions on the weights with respect to the true density ratio. By using that $p(x) = r_i(x) p_i^\prime(x)$, it follows for each $i \in \lbrace 1, \ldots, N^\prime \rbrace$ that
\begin{align*}
r(x) = \dfrac{r_i(x) p_i^\prime(x)}{\sum\limits_{j = 1}^{N^\prime} \omega_j p_j^\prime(x)}.
\end{align*}
So, for every $k \in \lbrace 1, \ldots, N^\prime \rbrace$ we have that
\begin{align*}
\sum\limits_{i = 1}^{N^\prime} \omega_i \mathbb{E}_{p_i^\prime(x)}[r(x)\phi(x)] &= \sum\limits_{i = 1}^{N^\prime} \omega_i \int r(x) \phi(x) p_i^\prime(x) dx = \int \phi(x) r(x) \left( \sum\limits_{i = 1}^{N^\prime} \omega_i p_i^\prime(x) \right) dx \\
&= \int \phi(x) r_k(x) p^\prime_k(x) dx = \mathbb{E}_{p^\prime_k(x)}[r_k(x) \phi(x)].
\end{align*}
Therefore, by summing over $k$, it follows that
\begin{align*}
\sum\limits_{i = 1}^{N^\prime} \omega_i \mathbb{E}_{p_i^\prime(x)}[r(x)\phi(x)] &= \sum_{k = 1}^{N^\prime} \left( \dfrac{1}{N^\prime} \right) \mathbb{E}_{p^\prime_k(x)}[r_k(x) \phi(x)],
\end{align*}
for which we can select the uniformly distributed weights $\Tilde{\omega}_i = \dfrac{1}{N^\prime}$ for every $i \in \lbrace 1, \ldots, N^\prime \rbrace$.\\
\vskip+0.25cm
$\bullet \,$ We end the present section with a brief remark about a particular case regarding the inference of \eqref{eq:5}. By using the form of the true density ratio, we obtain the following computations:
\begin{align*}
r(x) = \dfrac{p(x)}{p^\prime(x)} = \dfrac{p(x)}{\sum\limits_{i=1}^{N^\prime} \omega_i p_i^\prime(x)} = \dfrac{p(x)}{\sum\limits_{i=1}^{N^\prime-1} \omega_i p_i^\prime(x) + \omega_{N^\prime} p^\prime_{N^\prime}(x)}.     
\end{align*}
Let's consider the situation when $p^\prime_{N^\prime}(x) = p(x)$ and $\omega_{N^\prime} = \alpha \in [0, 1)$. Then, it follows that
\begin{align*}
r(x) = \dfrac{p(x)}{\sum\limits_{i=1}^{N^\prime-1} \omega_i p_i^\prime(x) + \alpha p(x)}, 
\end{align*}
where $\sum\limits_{i=1}^{N^\prime - 1} \omega_i + \alpha = \sum\limits_{i=1}^{N^\prime} \omega_i = 1$, so $\sum\limits_{i=1}^{N^\prime - 1} \omega_i = 1 - \alpha$. This can be considered as a generalization of the relative density ratio due to the fact that, when $N^\prime = 2$, one has access to the test dataset $\mathcal{X}$ and the train dataset $\mathcal{X}^\prime_1$, respectively. Therefore, one obtains the $\alpha$-relative density ratio $r_\alpha(x) = \dfrac{p(x)}{\alpha p(x) + (1-\alpha)p^\prime(x)}$. Finally, we highlight that, for $i \in \lbrace 1, \ldots, N^\prime \rbrace$, the sample sets $\mathcal{X}^\prime_i = \lbrace x^{\prime}_{j, (i)} \, | \, x^{\prime}_{j, (i)} \in \mathbb{R}^d \rbrace_{j = 1}^{n^\prime_i}$ corresponding to $p^\prime_i(x)$ form a partition of non-overlapping sets. Hence, the situation when $p^\prime_{N^\prime}(x) = p(x)$ is equivalent to the fact that $\mathcal{X}$ is non-overlapping with any other training subsets $\mathcal{X}^\prime_i$ for $i \in \lbrace 1, \ldots, N^\prime - 1 \rbrace$. In order to avoid this limitation, when dealing with the particular case of the generalized relative density ratio, we propose to formally use the same formulas as above despite the fact that the non-overlapping condition does not hold in general between train and test datasets, respectively.

\section{Proposed optimization method}\label{sec:G-KMM_ProposedOptMethod}
In what follows we consider our general KMM-type optimization technique based upon the computations made in the previous section. In the usual case of \eqref{KMM} and its extensions, one considers optimizing the MMD-based loss function involving the density ratio model only at the training points. Inspired by the techniques utilized in \eqref{RuLSIF} we propose to use a linear kernel model for the density ratio in order to obtain an optimization problem with respect to the underlying parameters of the model. It is worth pointing out that this methodology is similar to the one proposed in \cite{NeuralNetwork_KMM} where a neural network was used for the density ratio model. Furthermore, the training of the neural network model is made at each epoch with respect to non-overlapping shuffled data batches thus the setting from \cite{NeuralNetwork_KMM} is similar to ours (see also the alternative of the bootstrap aggregation technique from \cite{EfficientSamplingKMM}). But, the main difference is that, in our case, the train and test partitions are given at the beginning of the algorithm and are not randomly created at each iteration. \\
\vskip+0.25cm
Let's consider $N^\prime$ training sets $\mathcal{X}^\prime_l = \lbrace x^{\prime}_{j, (l)} \, | \, x^{\prime}_{j, (l)} \in \mathbb{R}^d \rbrace_{j = 1}^{n^\prime_l}$ such that $\mathcal{X}^\prime_{l} \overset{i.i.d.}{\sim} P^\prime_{l}$, where $P^\prime_{l}$ is the probability distribution with the underlying density $p^\prime_{l}$ for every $l \in \lbrace 1, \ldots, N^\prime \rbrace$. At the same time, let's suppose that we have $N$ test sets $\mathcal{X}_i = \lbrace x_{k,(i)} \, | \, x_{k, (i)} \in \mathbb{R}^d \rbrace_{k = 1}^{n_i}$ such that $\mathcal{X}_i \overset{i.i.d.}{\sim} P_{i}$ where $P_{i}$ is the probability distribution corresponding to the density $p_{i}$ for each $i \in \lbrace 1, \ldots, N \rbrace$. In what follows we will consider the test and train mixtures of probability densities $p(x) = \sum\limits_{i = 1}^{N} \omega_i p_i(x)$ and $p^\prime(x) = \sum\limits_{j = 1}^{N^\prime} \gamma_j p^\prime_j(x)$, where the weights satisfy $\sum\limits_{i = 1}^{N} \omega_i = \sum\limits_{j = 1}^{N^\prime} \gamma_j = 1$, with $\omega_i \in [0, 1]$ for every $i \in \lbrace 1, \ldots, N \rbrace$ and $\gamma_j \in [0, 1]$ for each $j \in \lbrace 1, \ldots, N^\prime \rbrace$. The general density ratio between the train and test samples is defined as 
\begin{align*}
r(x) = \dfrac{p(x)}{p^\prime(x)} = \dfrac{\sum\limits_{i = 1}^{N} \omega_i p_i(x)}{\sum\limits_{j = 1}^{N^\prime} \gamma_j p^\prime_j(x)}.    
\end{align*}
One observes that
\begin{align}\label{eq:6}
\mathbb{E}_{p(x)}[\phi(x)] &= \int \phi(x) p(x) dx = \int \phi(x) \left( \sum\limits_{i = 1}^{N} \omega_i p_i(x) \right) dx \nonumber \\
&= \sum\limits_{i = 1}^{N} \int \phi(x) \omega_i p_i(x) dx = \sum\limits_{i = 1}^{N} \omega_i \int \phi(x) p_i(x) dx = \sum\limits_{i = 1}^{N} \omega_i \mathbb{E}_{p_i(x)}[\phi(x)].
\end{align}
At the same time we have that
\begin{align}\label{eq:7}
\mathbb{E}_{p^\prime(x)}[r(x)\phi(x)] &= \int r(x) \phi(x) p^\prime(x) dx = \int r(x) \phi(x) \left( \sum\limits_{j = 1}^{N^\prime} \gamma_j p^\prime_j(x) \right) dx \nonumber\\
& = \sum\limits_{j = 1}^{N^\prime} \gamma_j \int r(x) \phi(x) p^\prime_j(x) dx = \sum\limits_{j = 1}^{N^\prime} \gamma_j \mathbb{E}_{p^\prime_j(x)}[r(x)\phi(x)].
\end{align}
Also
\begin{align}\label{eq:8}
\mathbb{E}_{p^\prime(x)}[r(x) \phi(x)] = \int r(x) \phi(x) p^\prime(x) dx = \int \phi(x) p(x) dx = \mathbb{E}_{p(x)}[\phi(x)] 
\end{align}
By combining \eqref{eq:6}, \eqref{eq:7} and \eqref{eq:8} we arrive at
\begin{align*}
\sum\limits_{j = 1}^{N^\prime} \gamma_j \mathbb{E}_{p^\prime_j(x)}[r(x)\phi(x)] = \sum\limits_{i = 1}^{N} \omega_i \mathbb{E}_{p_i(x)}[\phi(x)],    
\end{align*}
and taking into account that 
$$\mathbb{E}_{p^\prime(x)}[r(x)] = \int r(x) p^\prime(x) dx = \int p(x) dx = 1,$$ 
along with 
$$ \mathbb{E}_{p^\prime(x)}[r(x)] = \int r(x) p^\prime(x) dx = \int r(x) \left( \sum\limits_{j = 1}^{N^\prime} \gamma_j p^\prime_j(x) \right) dx = \sum\limits_{j = 1}^{N^\prime} \gamma_j \mathbb{E}_{p^\prime_j(x)} \left[ r(x) \right],$$
we therefore consider the following optimization problem:
\begin{align}\label{Generalized-KMM}\tag{OptPb-G-KMM}
\begin{cases}
&\min\limits_{\widehat{r}} \, \left\| \sum\limits_{j = 1}^{N^\prime} \gamma_j \mathbb{E}_{p^\prime_j(x)}[\hat{r}(x)\phi(x)] - \sum\limits_{i = 1}^{N} \omega_i \mathbb{E}_{p_i(x)}[\phi(x)] \right\|^2 \\ \\
&\text{subject to}
\begin{cases}
\hat{r}(x) \geq 0 \text{ for all } x \nonumber\\
\sum\limits_{j = 1}^{N^\prime} \gamma_j \mathbb{E}_{p^\prime_j(x)} \left[ \hat{r}(x) \right] = 1
\end{cases}
\end{cases}
\end{align}
where $\hat{r}(x)$ represents a density ratio model which approximates the true density ratio $r(x)$. In the following we shall show that the optimization problem presented in \eqref{Generalized-KMM} can be written as a \emph{quadratic optimization problem with constraints}. Then, the underlying loss function can be written as
\begin{align*}
\mathcal{L} := \left\| \sum\limits_{j = 1}^{N^\prime} \gamma_j \mathbb{E}_{p^\prime_j(x)}[\hat{r}(x)\phi(x)] - \sum\limits_{i = 1}^{N} \omega_i \mathbb{E}_{p_i(x)}[\phi(x)] \right\|^2 &= \left\| \sum\limits_{i = 1}^{N} \omega_i \mathbb{E}_{p_i(x)}[\phi(x)] \right\|^2 \\
&+ \left\langle \sum\limits_{j = 1}^{N^\prime} \gamma_j \mathbb{E}_{p^\prime_j(x)}[\hat{r}(x)\phi(x)], \sum\limits_{k = 1}^{N^\prime} \gamma_k \mathbb{E}_{p^\prime_k(x)}[\hat{r}(x)\phi(x)] \right\rangle \\
&- 2 \left\langle \sum\limits_{j = 1}^{N^\prime} \gamma_j \mathbb{E}_{p^\prime_j(x)}[\hat{r}(x)\phi(x)], \sum\limits_{i = 1}^{N} \omega_i \mathbb{E}_{p_i(x)}[\phi(x)] \right\rangle.
\end{align*}
Ignoring constants irrelevant with respect to $\hat{r}(x)$, the objective function defined above can be taken as
\begin{align*}
\mathcal{L} &= \left\langle \sum\limits_{j = 1}^{N^\prime} \gamma_j \mathbb{E}_{p^\prime_j(x)}[\hat{r}(x)\phi(x)], \sum\limits_{k = 1}^{N^\prime} \gamma_k \mathbb{E}_{p^\prime_k(x)}[\hat{r}(x)\phi(x)] \right\rangle - 2 \left\langle \sum\limits_{j = 1}^{N^\prime} \gamma_j \mathbb{E}_{p^\prime_j(x)}[\hat{r}(x)\phi(x)], \sum\limits_{i = 1}^{N} \omega_i \mathbb{E}_{p_i(x)}[\phi(x)] \right\rangle,
\end{align*}
hence
\begin{align*}
\mathcal{L} = \sum\limits_{j = 1}^{N^\prime}\sum\limits_{k = 1}^{N^\prime} \left\langle \gamma_j \mathbb{E}_{p^\prime_j(x)}[\hat{r}(x)\phi(x)], \gamma_k \mathbb{E}_{p^\prime_k(x)}[\hat{r}(x)\phi(x)] \right\rangle - 2 \sum\limits_{j = 1}^{N^\prime}\sum\limits_{i = 1}^{N} \left\langle \gamma_j \mathbb{E}_{p^\prime_j(x)}[\hat{r}(x)\phi(x)], \omega_i \mathbb{E}_{p_i(x)}[\phi(x)] \right\rangle.    
\end{align*}
By consider employing empirical averages, we therefore obtain that $\mathbb{E}_{p_i(x)}[\phi(x)] \approx \dfrac{1}{n_i} \sum\limits_{l = 1}^{n_i} \phi(x_{l, (i)})$ and $\mathbb{E}_{p^\prime_j(x)}[\hat{r}(x)\phi(x)] \approx \dfrac{1}{n^\prime_j} \sum\limits_{t = 1}^{n^\prime_j} \hat{r} \left( x^\prime_{t, (j)} \right) \phi \left( x^\prime_{t, (j)} \right)$, which implies that the empirical loss function $\widehat{\mathcal{L}}$ which approximates $\mathcal{L}$ takes the form
\begin{align*}
\widehat{\mathcal{L}} &= \sum\limits_{j = 1}^{N^\prime}\sum\limits_{k = 1}^{N^\prime} \left\langle \gamma_j \left( \dfrac{1}{n^\prime_j} \sum\limits_{t = 1}^{n^\prime_j} \hat{r} \left( x^\prime_{t, (j)} \right) \phi \left( x^\prime_{t, (j)} \right) \right), \gamma_k \left( \dfrac{1}{n^\prime_k} \sum\limits_{s = 1}^{n^\prime_k} \hat{r} \left( x^\prime_{s, (k)} \right) \phi \left( x^\prime_{s, (k)} \right) \right) \right\rangle \\
&- 2 \sum\limits_{j = 1}^{N^\prime}\sum\limits_{i = 1}^{N} \left\langle \gamma_j \left( \dfrac{1}{n^\prime_j} \sum\limits_{t = 1}^{n^\prime_j} \hat{r} \left( x^\prime_{t, (j)} \right) \phi \left( x^\prime_{t, (j)} \right) \right), \omega_i \left( \dfrac{1}{n_i} \sum\limits_{l = 1}^{n_i} \phi(x_{l, (i)}) \right) \right\rangle.   
\end{align*}
Taking $n^\prime_{max} := \max\limits_{j \in \lbrace 1, \ldots, N^\prime \rbrace} \lbrace n^\prime_j \rbrace$ and multiplying $\widehat{\mathcal{L}}$ with $\dfrac{1}{2} \left( n^\prime_{max} \right)^2$, we simplify the previous identity as follows:
\begin{align*}
\widehat{\mathcal{L}} &= \sum\limits_{j = 1}^{N^\prime}\sum\limits_{k = 1}^{N^\prime} \left( \dfrac{\left( n^\prime_{max} \right)^2}{n^\prime_j n^\prime_k} \dfrac{\gamma_j \gamma_k}{2} \left\langle \sum\limits_{t = 1}^{n^\prime_j} \hat{r} \left( x^\prime_{t, (j)} \right) \phi \left( x^\prime_{t, (j)} \right), \sum\limits_{s = 1}^{n^\prime_k} \hat{r} \left( x^\prime_{s, (k)} \right) \phi \left( x^\prime_{s, (k)} \right) \right\rangle \right) \\
& - \sum\limits_{j = 1}^{N^\prime}\sum\limits_{i = 1}^{N} \left( \dfrac{\left( n^\prime_{max} \right)^2}{n_i n^\prime_j} \gamma_j \omega_i \left\langle \sum\limits_{t = 1}^{n^\prime_j} \hat{r} \left( x^\prime_{t, (j)} \right) \phi \left( x^\prime_{t, (j)} \right), \sum\limits_{l = 1}^{n_i} \phi \left( x_{l, (i)} \right) \right\rangle \right).
\end{align*}
By utilizing the linearity of the inner product, we obtain
\begin{align*}
\widehat{\mathcal{L}} &= \sum\limits_{j = 1}^{N^\prime}\sum\limits_{k = 1}^{N^\prime} \left( \dfrac{\left( n^\prime_{max} \right)^2}{n^\prime_j n^\prime_k} \dfrac{\gamma_j \gamma_k}{2} \sum\limits_{t = 1}^{n^\prime_j} \sum\limits_{s = 1}^{n^\prime_k} \hat{r} \left( x^\prime_{t, (j)} \right) \left\langle \phi \left( x^\prime_{t, (j)} \right), \phi \left( x^\prime_{s, (k)} \right) \right\rangle \hat{r} \left( x^\prime_{s, (k)} \right) \right) \\
& - \sum\limits_{j = 1}^{N^\prime}\sum\limits_{i = 1}^{N} \left( \dfrac{\left( n^\prime_{max} \right)^2}{n_i n^\prime_j} \gamma_j \omega_i \sum\limits_{t = 1}^{n^\prime_j} \sum\limits_{l = 1}^{n_i} \hat{r} \left( x^\prime_{t, (j)} \right) \left\langle \phi \left( x^\prime_{t, (j)} \right), \phi \left( x_{l, (i)} \right) \right\rangle \right).
\end{align*}
From the definition of the kernel mapping as an inner product of the feature maps i.e., $K(x, y) = \langle \phi(x), \phi(y) \rangle$, it follows that
\begin{align}\label{eq:9}
\widehat{\mathcal{L}} &= \sum\limits_{j = 1}^{N^\prime}\sum\limits_{k = 1}^{N^\prime} \left( \dfrac{\left( n^\prime_{max} \right)^2}{n^\prime_j n^\prime_k} \dfrac{\gamma_j \gamma_k}{2} \sum\limits_{t = 1}^{n^\prime_j} \sum\limits_{s = 1}^{n^\prime_k} \hat{r} \left( x^\prime_{t, (j)} \right) K \left( x^\prime_{t, (j)}, x^\prime_{s, (k)} \right) \hat{r} \left( x^\prime_{s, (k)} \right) \right) \nonumber \\
& - \sum\limits_{j = 1}^{N^\prime}\sum\limits_{i = 1}^{N} \left( \dfrac{\left( n^\prime_{max} \right)^2}{n_i n^\prime_j} \gamma_j \omega_i \sum\limits_{t = 1}^{n^\prime_j} \sum\limits_{l = 1}^{n_i} \hat{r} \left( x^\prime_{t, (j)} \right) K \left( x^\prime_{t, (j)}, x_{l, (i)} \right) \right).
\end{align}
In order to simplify the previous computations we consider the following notations:
\begin{align*}
&\hat{r}_{\mathcal{X}^\prime_j} := \left( \hat{r} \left( x^\prime_{1, (j)} \right), \ldots, \hat{r} \left( x^\prime_{n^\prime_j, (j)} \right) \right)^T \in \mathbb{R}^{n^\prime_j \times 1} \, ,\, j \in \lbrace 1, \ldots, N^\prime \rbrace \\
&\hat{r}_{\mathcal{X}_i} := \left( \hat{r} \left( x_{1, (i)} \right), \ldots, \hat{r} \left( x_{n_i, (i)} \right) \right)^T \in \mathbb{R}^{n_i \times 1} \, ,\, i \in \lbrace 1, \ldots, N \rbrace.
\end{align*}
At the same time we define for $i \in \lbrace 1, \ldots, N \rbrace$ and $j \in \lbrace 1, \ldots, N^\prime \rbrace$ the vector $h^{[i, j]} \in \mathbb{R}^{n^\prime_j \times 1}$, such that
\begin{align*}
&h_t^{[i, j]} = \dfrac{\left( n^\prime_{max} \right)^2}{n_i n^\prime_j} \gamma_j \omega_i \sum\limits_{l = 1}^{n_i} K \left( x^\prime_{t, (j)}, x_{l, (i)} \right) \text{ for each } t \in \lbrace 1, \ldots, n^\prime_j \rbrace.    
\end{align*}
Also, for every $j, k \in \lbrace 1, \ldots, N^\prime \rbrace$ we define the matrix $H^{[j, k]} \in \mathbb{R}^{n^\prime_j \times n^\prime_k}$, such that
\begin{align*}
H^{[j, k]}_{t, s} = \dfrac{\left( n^\prime_{max} \right)^2}{n^\prime_j n^\prime_k} \dfrac{\gamma_j \gamma_k}{2} K \left( x^\prime_{t, (j)}, x^\prime_{s, (k)} \right) \text{ for each } t \in \lbrace 1, \ldots, n^\prime_j \rbrace \text{ and } s \in \lbrace 1, \ldots, n^\prime_k \rbrace.    
\end{align*}
By using the above notations we obtain the following computations:
\begin{align}\label{eq:10}
\dfrac{\left( n^\prime_{max} \right)^2}{n_i n^\prime_j} \gamma_j \omega_i \sum\limits_{t = 1}^{n^\prime_j} \sum\limits_{l = 1}^{n_i} \hat{r} \left( x^\prime_{t, (j)} \right) K \left( x^\prime_{t, (j)}, x_{l, (i)} \right) &= \sum\limits_{t = 1}^{n^\prime_j} \hat{r} \left( x^\prime_{t, (j)} \right) \left( \dfrac{\left( n^\prime_{max} \right)^2}{n_i n^\prime_j} \gamma_j \omega_i \sum\limits_{l = 1}^{n_i} K \left( x^\prime_{t, (j)}, x_{l, (i)} \right) \right) \nonumber \\
&= \sum\limits_{t = 1}^{n^\prime_j} \hat{r} \left( x^\prime_{t, (j)} \right) h_t^{[i, j]} \nonumber\\
&= \left( h^{[i, j]} \right)^T \hat{r}_{\mathcal{X}^\prime_j}.
\end{align}
On the other hand, by denoting
\begin{align}\label{eq:11}
C^{[j, k]} := \dfrac{\left( n^\prime_{max} \right)^2}{n^\prime_j n^\prime_k} \dfrac{\gamma_j \gamma_k}{2} \sum\limits_{t = 1}^{n^\prime_j} \sum\limits_{s = 1}^{n^\prime_k} \hat{r} \left( x^\prime_{t, (j)} \right) K \left( x^\prime_{t, (j)}, x^\prime_{s, (k)} \right) \hat{r} \left( x^\prime_{s, (k)} \right),    
\end{align}
we find that
\begin{align}\label{eq:12}
C^{[j, k]} &= \sum\limits_{t = 1}^{n^\prime_j} \sum\limits_{s = 1}^{n^\prime_k} \hat{r} \left( x^\prime_{t, (j)} \right) \left( \dfrac{\left( n^\prime_{max} \right)^2}{n^\prime_j n^\prime_k} \dfrac{\gamma_j \gamma_k}{2} K \left( x^\prime_{t, (j)}, x^\prime_{s, (k)} \right) \right) \hat{r} \left( x^\prime_{s, (k)} \right) \nonumber \\
&= \sum\limits_{t = 1}^{n^\prime_j} \hat{r} \left( x^\prime_{t, (j)} \right) \sum\limits_{s = 1}^{n^\prime_k} \left( \dfrac{\left( n^\prime_{max} \right)^2}{n^\prime_j n^\prime_k} \dfrac{\gamma_j \gamma_k}{2} K \left( x^\prime_{t, (j)}, x^\prime_{s, (k)} \right) \hat{r} \left( x^\prime_{s, (k)} \right) \right) \nonumber \\
&= \sum\limits_{t = 1}^{n^\prime_j} \hat{r} \left( x^\prime_{t, (j)} \right) \sum\limits_{s = 1}^{n^\prime_k} H^{[j, k]}_{t, s} \hat{r} \left( x^\prime_{s, (k)} \right) \nonumber \\
&= \sum\limits_{t = 1}^{n^\prime_j} \hat{r} \left( x^\prime_{t, (j)} \right) \left( H^{[j, k]} \hat{r}_{\mathcal{X}^\prime_k} \right)_t \nonumber \\
&= \left( \hat{r}_{\mathcal{X}^\prime_j} \right)^T H^{[j, k]} \left( \hat{r}_{\mathcal{X}^\prime_k} \right).
\end{align}
By merging \eqref{eq:9}, \eqref{eq:10}, \eqref{eq:11} and \eqref{eq:12}, we find a simpler formulation of the empirical loss, namely
\begin{align}\label{eq:13}
\widehat{\mathcal{L}} &= \sum\limits_{j = 1}^{N^\prime}\sum\limits_{k = 1}^{N^\prime} \left( \left( \hat{r}_{\mathcal{X}^\prime_j} \right)^T H^{[j, k]} \left( \hat{r}_{\mathcal{X}^\prime_k} \right) \right) - \sum\limits_{j = 1}^{N^\prime}\sum\limits_{i = 1}^{N} \left( \left( h^{[i, j]} \right)^T \left( \hat{r}_{\mathcal{X}^\prime_j} \right) \right).   
\end{align}
In order to make our method more similar to RuLSIF we consider modeling the true density ratio as a linear model i.e., $\hat{r}(x) = \langle \theta, \xi(x) \rangle$ where $\theta = (\theta_1, \ldots, \theta_b) \in \mathbb{R}^b$ and $\xi: \mathbb{R}^d \to \mathbb{R}^b$ such that $\xi(x) = (\xi_1(x), \ldots, \xi_b(x))$ for each $x \in \mathbb{R}^d$. For $j \in \lbrace 1, \ldots, N^\prime \rbrace$ we define the matrix $A^{[j]} \in \mathbb{R}^{n^\prime_j \times b}$ such that
\begin{align}\label{eq:14}
A^{[j]} = 
\begin{pmatrix}
\xi_1( x^\prime_{1, (j)} ) & \ldots & \xi_b ( x^\prime_{1, (j)} ) \\ \\
\vdots & \vdots & \vdots \\ \\
\xi_1( x^\prime_{n^\prime_j, (j)} ) & \ldots & \xi_b ( x^\prime_{n^\prime_j, (j)} )
\end{pmatrix}
\end{align}
Some easy computations reveal that
\begin{align}\label{eq:15}
A^{[j]} \theta = 
\begin{pmatrix}
\xi^T ( x^\prime_{1, (j)} ) \theta \\
\ldots \\
\xi^T ( x^\prime_{n^\prime_j, (j)} ) \theta
\end{pmatrix} = \hat{r}_{\mathcal{X}^\prime_j} \in \mathbb{R}^{n^\prime_j \times 1}
\end{align}
Therefore, \eqref{eq:15} implies that
\begin{align}\label{eq:16}
\left( \hat{r}_{\mathcal{X}^\prime_j} \right)^T H^{[j, k]} \left( \hat{r}_{\mathcal{X}^\prime_k} \right) = (A^{[j]} \theta)^T H^{[j, k]} (A^{[k]} \theta) = \theta^T \left( (A^{[j]})^T H^{[j, k]} A^{[k]} \right) \theta.    
\end{align}
Using again \eqref{eq:15}, it follows that
\begin{align}\label{eq:17}
\left( h^{[i, j]} \right)^T \left( \hat{r}_{\mathcal{X}^\prime_j} \right) = \left( h^{[i, j]} \right)^T A^{[j]} \theta = \left( \left( h^{[i, j]} \right)^T A^{[j]} \right) \theta.
\end{align}
Consequently, \eqref{eq:13}, \eqref{eq:16} and \eqref{eq:17} imply that
\begin{align}\label{OptPb-Quadratic-KMM}
\widehat{\mathcal{L}} &= \theta^T \left( \sum\limits_{j = 1}^{N^\prime}\sum\limits_{k = 1}^{N^\prime} (A^{[j]})^T H^{[j, k]} A^{[k]} \right) \theta - \left( \sum\limits_{j = 1}^{N^\prime}\sum\limits_{i = 1}^{N} (h^{[i, j]})^T A^{[j]} \right) \theta. 
\end{align}
For the particular case of kernel methods we can choose $\xi(x)$ with the same technique as in the case of RuLSIF, namely we will use the test dataset defined as the reunion of the non-overlapping test sample datasets: $\mathcal{X} = \bigcup\limits_{i = 1}^{N} \mathcal{X}_i$. Therefore, we will select $\xi_k(x) = K(x, x_k)$ where $x_k \in \mathcal{X}$ for each $k \in \lbrace 1, \ldots, b \rbrace$, thus $b = \sum\limits_{i = 1}^{N} n_i$. \\
In what follows we will select $K$ as a kernel endowed with non-negative values, such as the \emph{RBF kernel} or the \emph{Laplacian kernel}. In \cite{Gretton_KMM} the value of $\hat{r}(x)$ was bounded (only with respect to the training samples) in the interval $[0, B]$, where $B > 0$. In our case, in order to simplify this condition, we consider $\hat{r}(x) = \langle \theta, \xi(x) \rangle \geq 0$ and using that $K$ takes non-negative values, we shall impose that $\theta_k \in [0, B]$ for every $k \in \lbrace 1, \ldots, b \rbrace$, where $B > 0$ is a constant chosen up to our choice. On the other hand, we define $\Xi = (\Xi_1, \ldots, \Xi_b) \in \mathbb{R}^b$ as
$$\Xi := \sum\limits_{j = 1}^{N^\prime} \left( \dfrac{\gamma_j}{n^\prime_j} \right) \sum\limits_{k = 1}^{n^\prime_j} \xi(x_{k, (j)}).$$
The constraint $\sum\limits_{j = 1}^{N^\prime} \gamma_j \mathbb{E}_{p^\prime_j(x)} \left[ \hat{r}(x) \right] = 1$ from \eqref{Generalized-KMM} leads to its empirical counterpart, namely
\begin{align*}
1 \approx \sum\limits_{j = 1}^{N^\prime} \gamma_j \left( \dfrac{1}{n^\prime_j} \sum\limits_{k = 1}^{n^\prime_j} \hat{r}(x_{k, (j)}) \right) = \sum\limits_{j = 1}^{N^\prime} \left( \dfrac{\gamma_j}{n^\prime_j} \right) \sum\limits_{k = 1}^{n^\prime_j} \hat{r}(x_{k, (j)}) = \sum\limits_{j = 1}^{N^\prime} \left( \dfrac{\gamma_j}{n^\prime_j} \right) \sum\limits_{k = 1}^{n^\prime_j} \langle \theta, \xi(x_{k, (j)}) \rangle = \langle \theta, \Xi \rangle.
\end{align*}
Using the above identity $\langle \theta, \Xi \rangle = 1$, similar to the numerical description of KMM from \cite{Gretton_KMM}, we consider $\varepsilon > 0$ such that $| \langle \theta, \Xi \rangle - 1 | \leq \varepsilon$, hence $\sum\limits_{k = 1}^{b} \theta_k \Xi_k \leq \varepsilon + 1$ and $- \sum\limits_{k = 1}^{b} \theta_k \Xi_k \leq \varepsilon - 1 $, respectively. \\
By combining \eqref{OptPb-Quadratic-KMM} with the constraints presented above, we finally obtain our \texttt{Generalized KMM} method represented by the following \emph{empirical generalized KMM optimization problem}:
\begin{align}\label{Empirical-Generalized-KMM}\tag{OptPb-Empirical-G-KMM}
\begin{cases}
&\min\limits_{\theta} \, \left[ \theta^T \left( \sum\limits_{j = 1}^{N^\prime}\sum\limits_{k = 1}^{N^\prime} (A^{[j]})^T H^{[j, k]} A^{[k]} \right) \theta - \left( \sum\limits_{j = 1}^{N^\prime}\sum\limits_{i = 1}^{N} (h^{[i, j]})^T A^{[j]} \right) \theta \right] \nonumber\\ \\
&\text{subject to}
\begin{cases}
\theta_k \in [0, B] \text{ for } k \in \lbrace 1, \ldots, b \rbrace \nonumber \\
+ \sum\limits_{k = 1}^{b} \theta_k \Xi_k \leq \varepsilon + 1 \\
- \sum\limits_{k = 1}^{b} \theta_k \Xi_k \leq \varepsilon - 1 \nonumber.
\end{cases}
\end{cases}
\end{align}

\section{Experiments}\label{sec:G-KMM_Experiments}
\begin{figure}[!ht]
\centering
\includegraphics[width=1.0\textwidth]{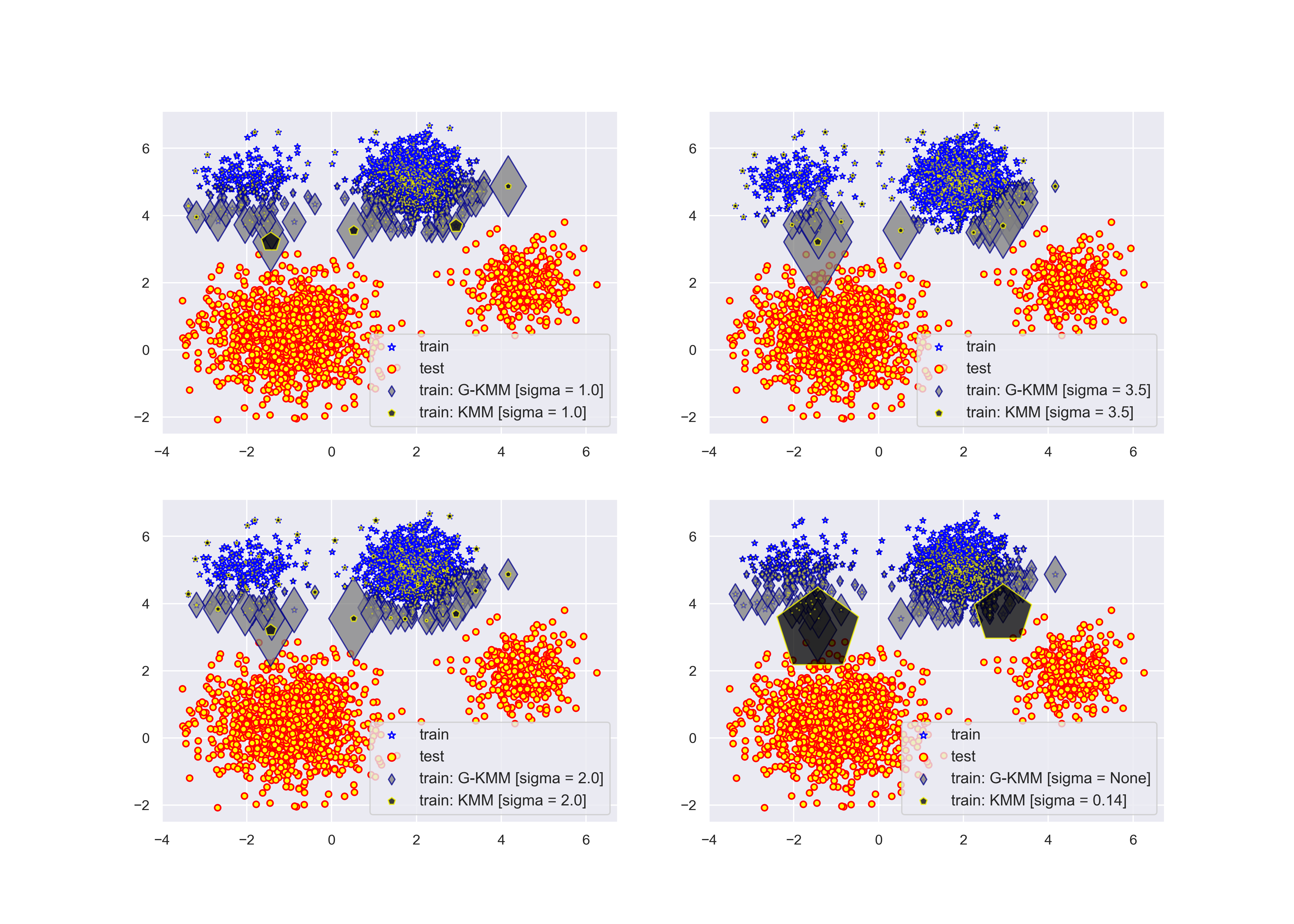}
\caption{KMM vs. \texttt{Generalized KMM}}
\label{fig:1-GKMM-clusters}
\end{figure}

In this section we present some numerical simulations based on our implementation of the \texttt{Generalized KMM} optimizer concerning certain experiments made on some synthetic datasets. We highlight that our codes hinge on \texttt{SKLearn} \cite{scikit-learn} and the \texttt{CVXPY} package \cite{CVXPY_2, CVXPY_1}, respectively. At the same time, all the details about our implementation and the corresponding experiments can be found in our \texttt{GitHub} link presented in the \emph{Abstract} of the present paper. It is of utmost importance to mention that our parameter $\sigma$ which will appear in the experiments from the following sequel and in the underlying implementation, is denoted as $\gamma$ in the \texttt{SKLearn} implementation (and when $\gamma \approx \tilde{\sigma}^{-2}$ then the aforementioned parameter is associated with the variance $\tilde{\sigma}^2$ of the kernel $K$).\\ \\
Our first experiment is related to an application of the \texttt{Generalized KMM} described through the optimization problem \eqref{Empirical-Generalized-KMM},  along with the classical KMM method, respectively. The implementation for the underlying KMM algorithm is inspired by the codes belonging to \cite{ADAPT} and \cite{DIW}, respectively. For this experiment we have considered $4$ clusters (two of them belonging to the training dataset and the other two representing the test samples) consisting of different number of samples, i.e. $200$, $1000$, $1000$ and $300$, respectively. The clusters were generated using the function \texttt{make$\_$blobs} from \texttt{SKLearn} with the following standard deviation values: $0.6$, $0.6$, $0.9$ and $0.6$, respectively. For both numerical methods, the parameter $B$ was set to $1000$ while the values of the parameter $\sigma$ were chosen as $1.0$, $3.5$, $2.0$ and $None$. We mention that $None$ is equivalent to the default value used in the definition of the kernels from \texttt{SKLearn}. In the case of the \texttt{Generalized KMM} algorithm each vector and matrix that is defined through a kernel has the same default value defined as \texttt{1/n\_features}, but we have chosen to write it as $None$ since this value is already shown in the plots related to the KMM method (due to the fact that we use the same datasets for both methods hence we have the same number of features). From the results depicted in figure \eqref{fig:1-GKMM-clusters} we observe that the classical KMM method has weight values smaller than those of the \texttt{Generalized KMM} algorithm. Our optimization method gives weights a higher value near the boundary of the two training clusters, while KMM emphasize also the samples belonging near the center of the training data. One also observers that by increasing the $\sigma$ parameter the values of the weights also increase. On the other hand, the particular case when $\sigma$ is attributed the value of $None$ (depicted in the bottom right plot) shows that KMM leads to fewer weights with high values in contrast with the \texttt{Generalized KMM} method.

\vskip-0.35cm

\begin{figure}[!ht]
\centering
    \begin{subfigure}[b]{0.44\textwidth}
        \includegraphics[scale=0.23]{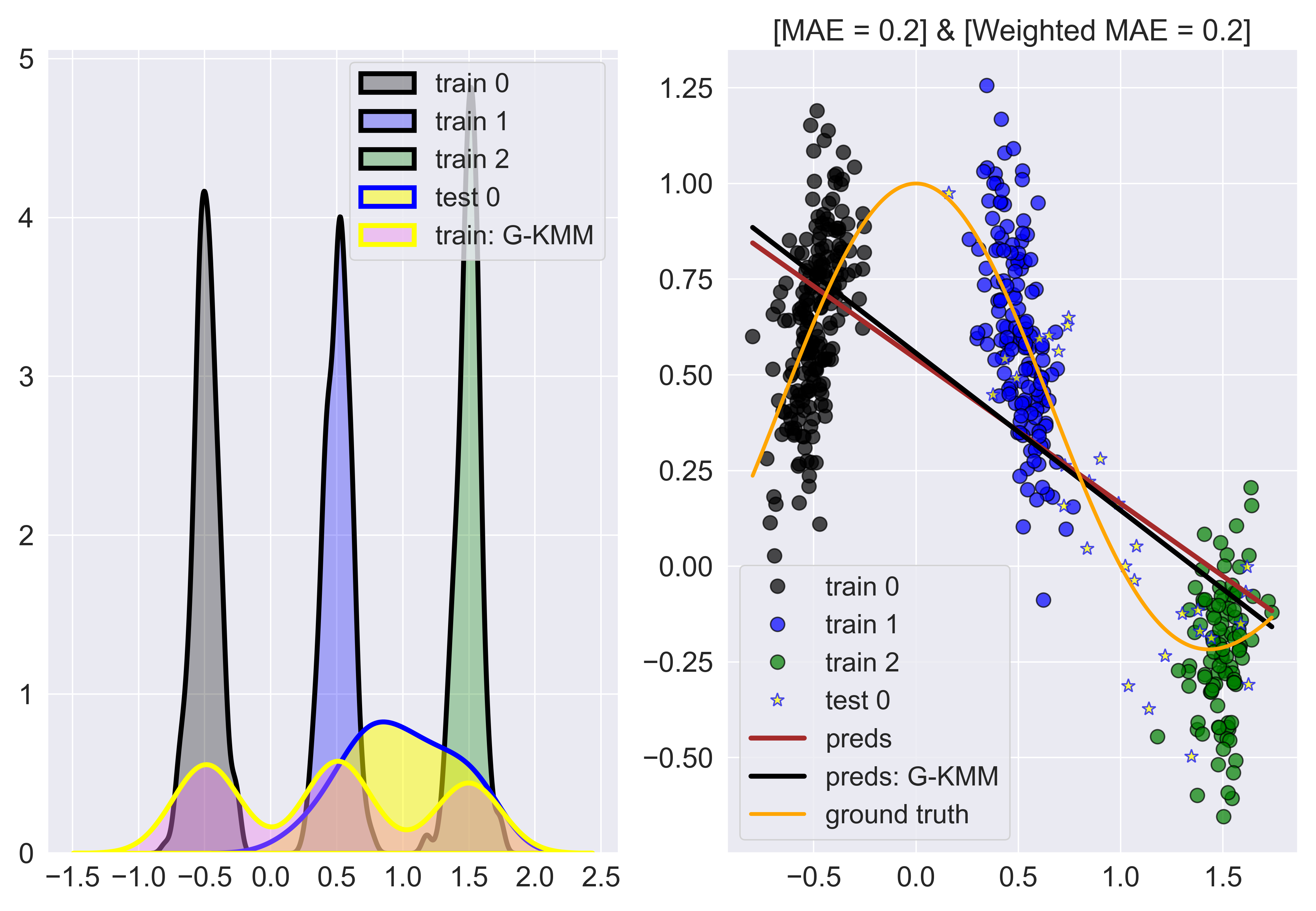}
    \end{subfigure}
    \begin{subfigure}[b]{0.44\textwidth}
        \includegraphics[scale=0.23]{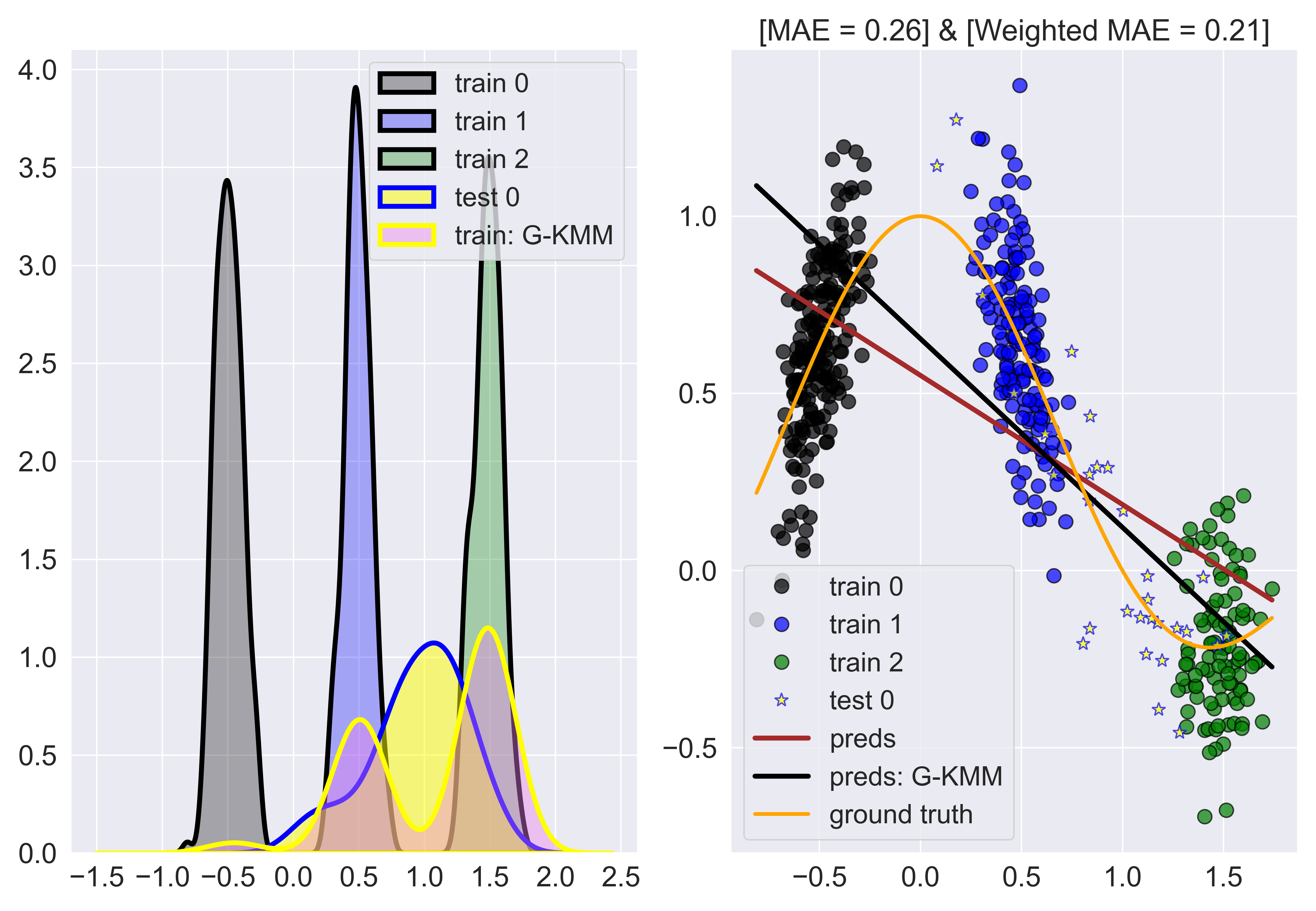}
    \end{subfigure}
    
    \begin{subfigure}[b]{0.44\textwidth}
        \includegraphics[scale=0.23]{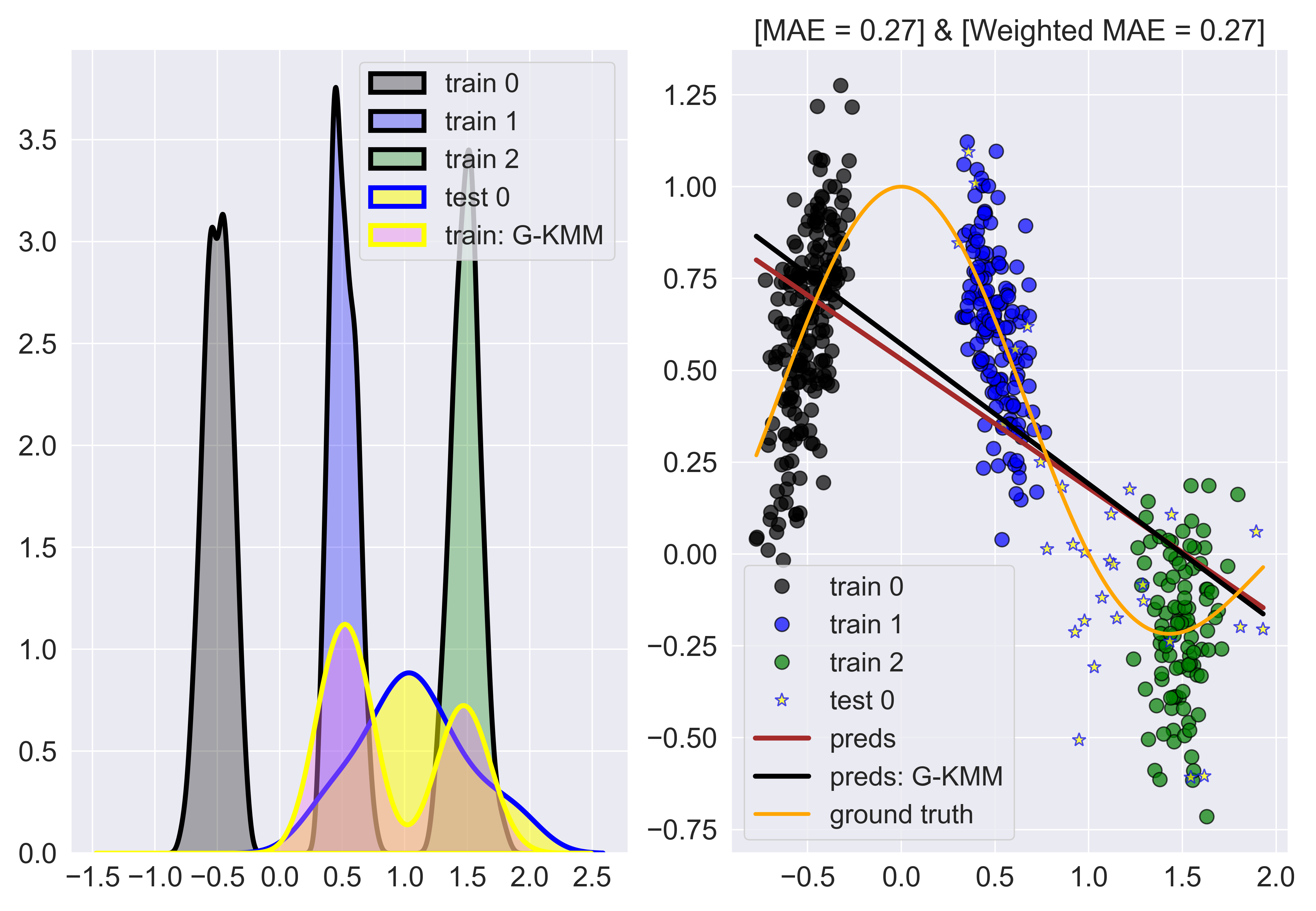}
    \end{subfigure}
    
\caption{Multiple train datasets}
\label{fig:2-GKMM-regression-train}
\end{figure}

\begin{figure}[!ht]
\centering
    \begin{subfigure}[b]{0.44\textwidth}
        \includegraphics[scale=0.23]{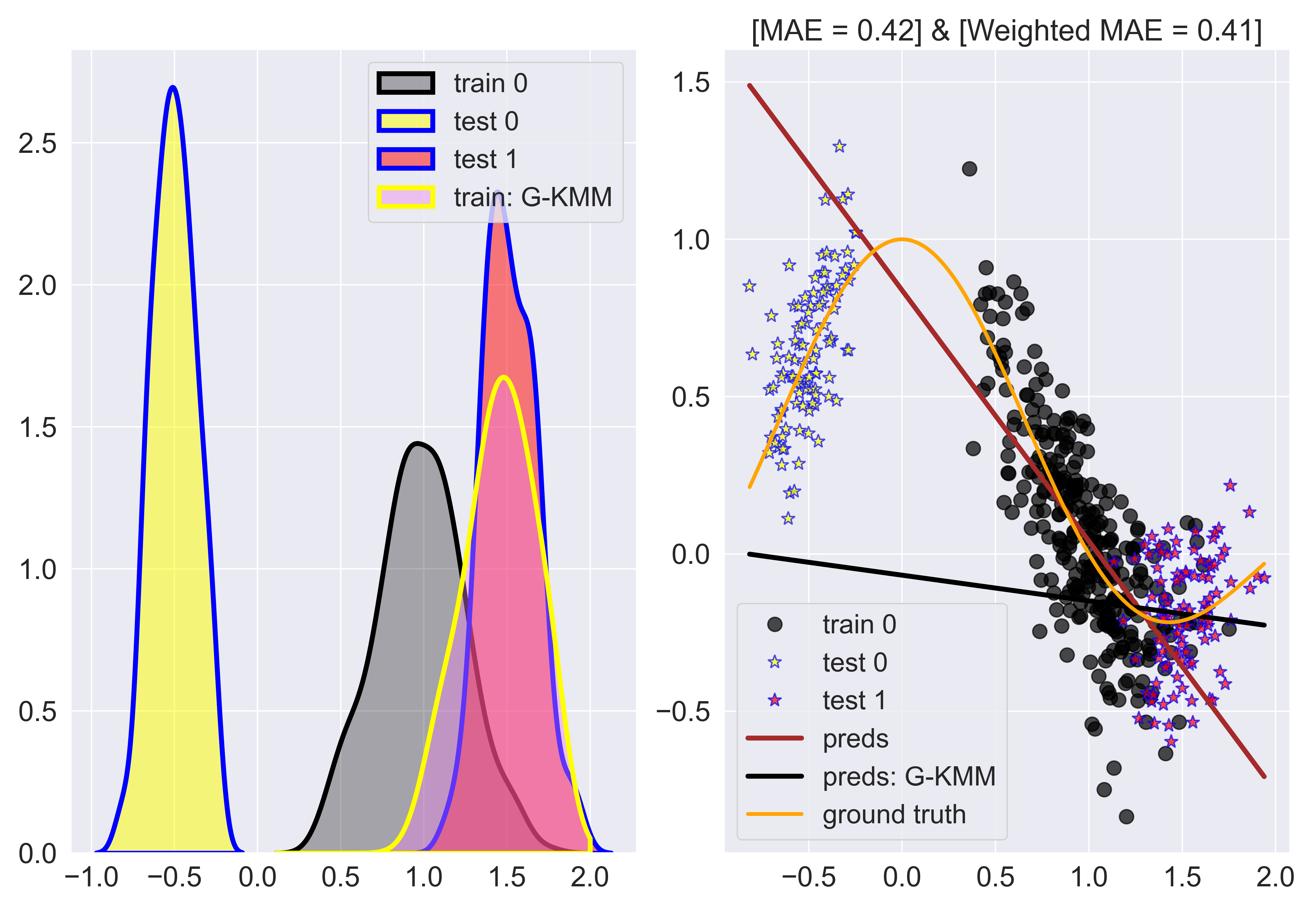}
    \end{subfigure}
    \begin{subfigure}[b]{0.44\textwidth}
        \includegraphics[scale=0.23]{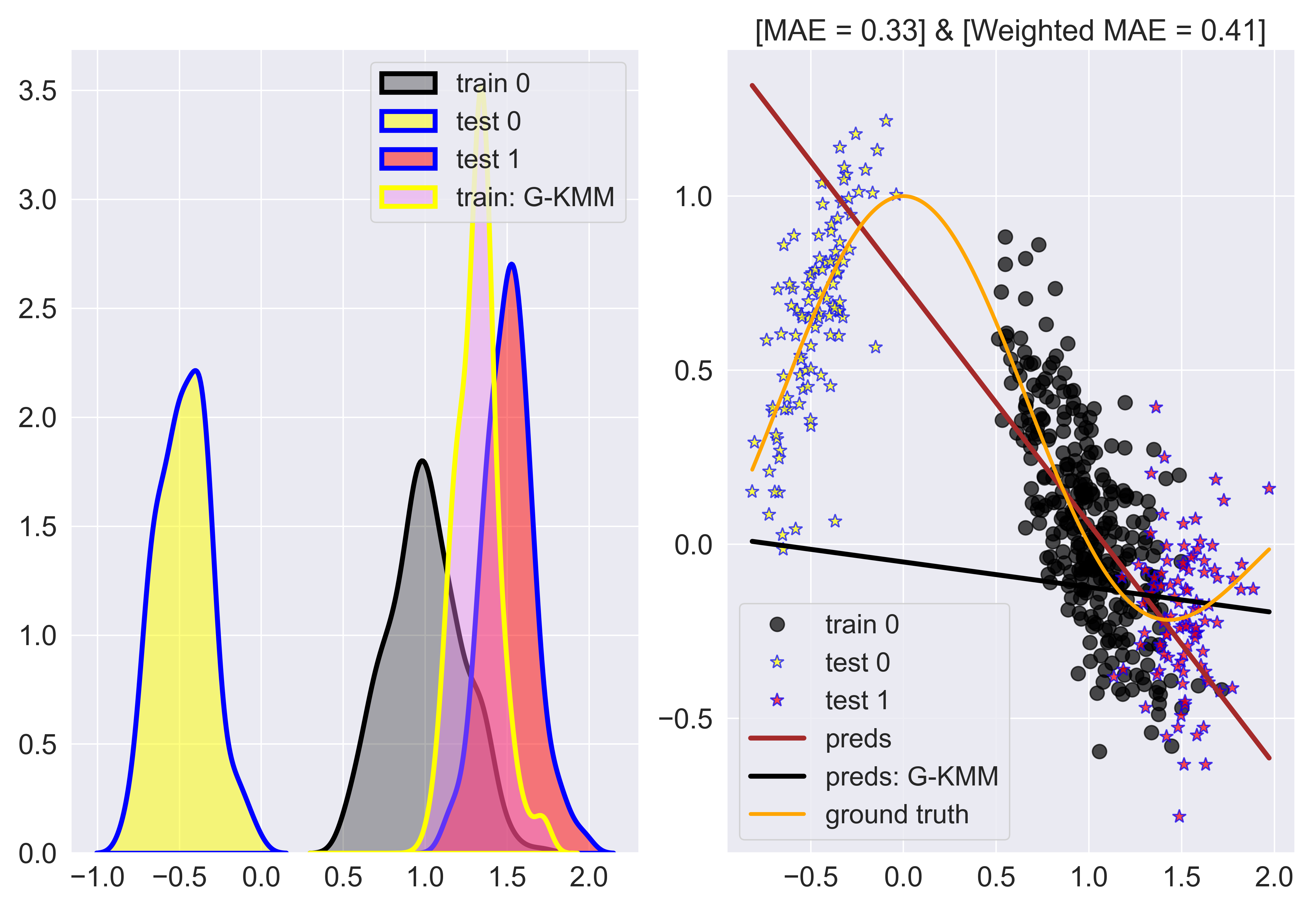}
    \end{subfigure}
    
    \begin{subfigure}[b]{0.44\textwidth}
        \includegraphics[scale=0.23]{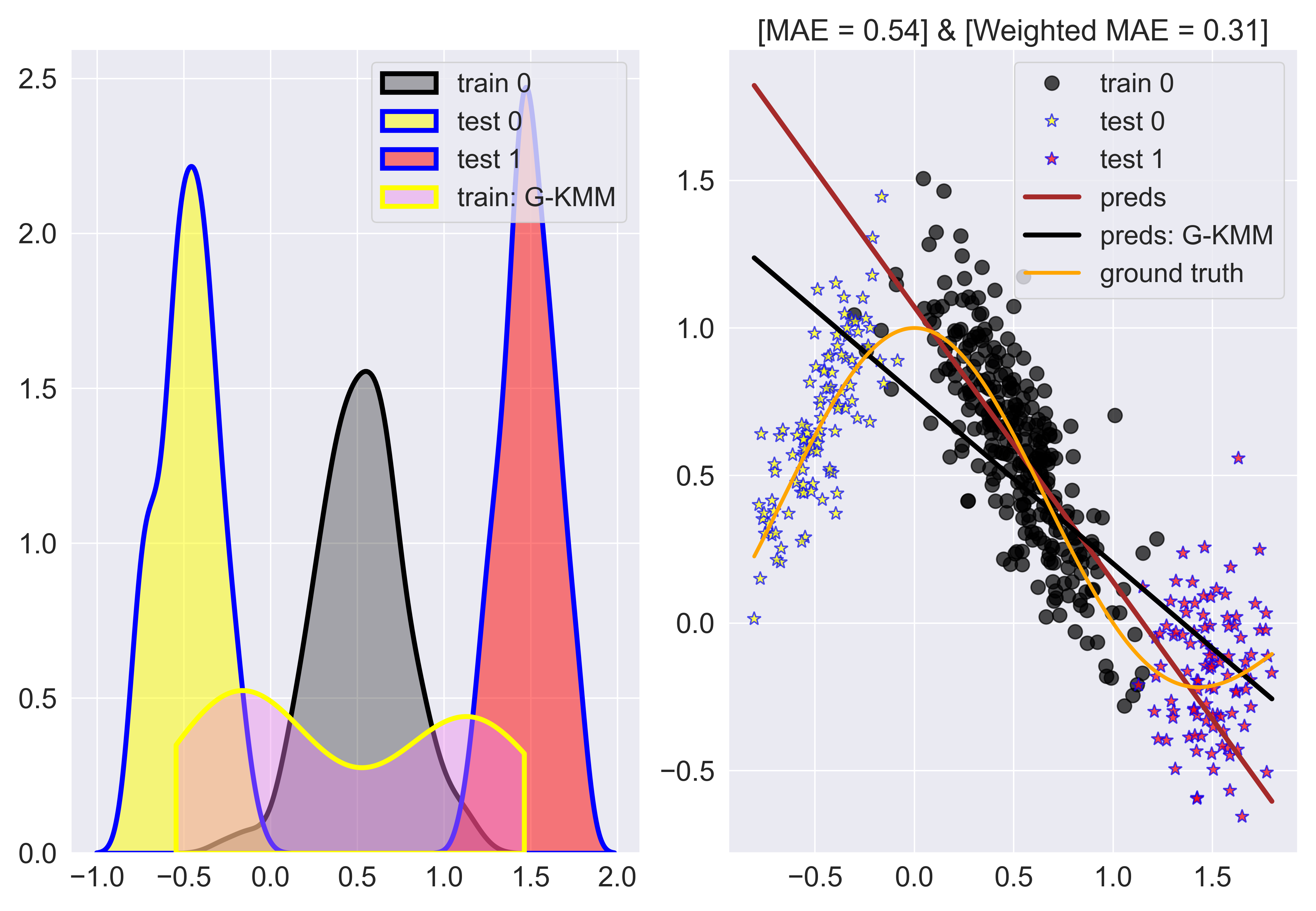}
    \end{subfigure}
    
\caption{Multiple test datasets}
\label{fig:3-GKMM-regression-test}
\end{figure}

\begin{figure}[!ht]
\centering
    \begin{subfigure}[b]{0.44\textwidth}
        \includegraphics[scale=0.23]{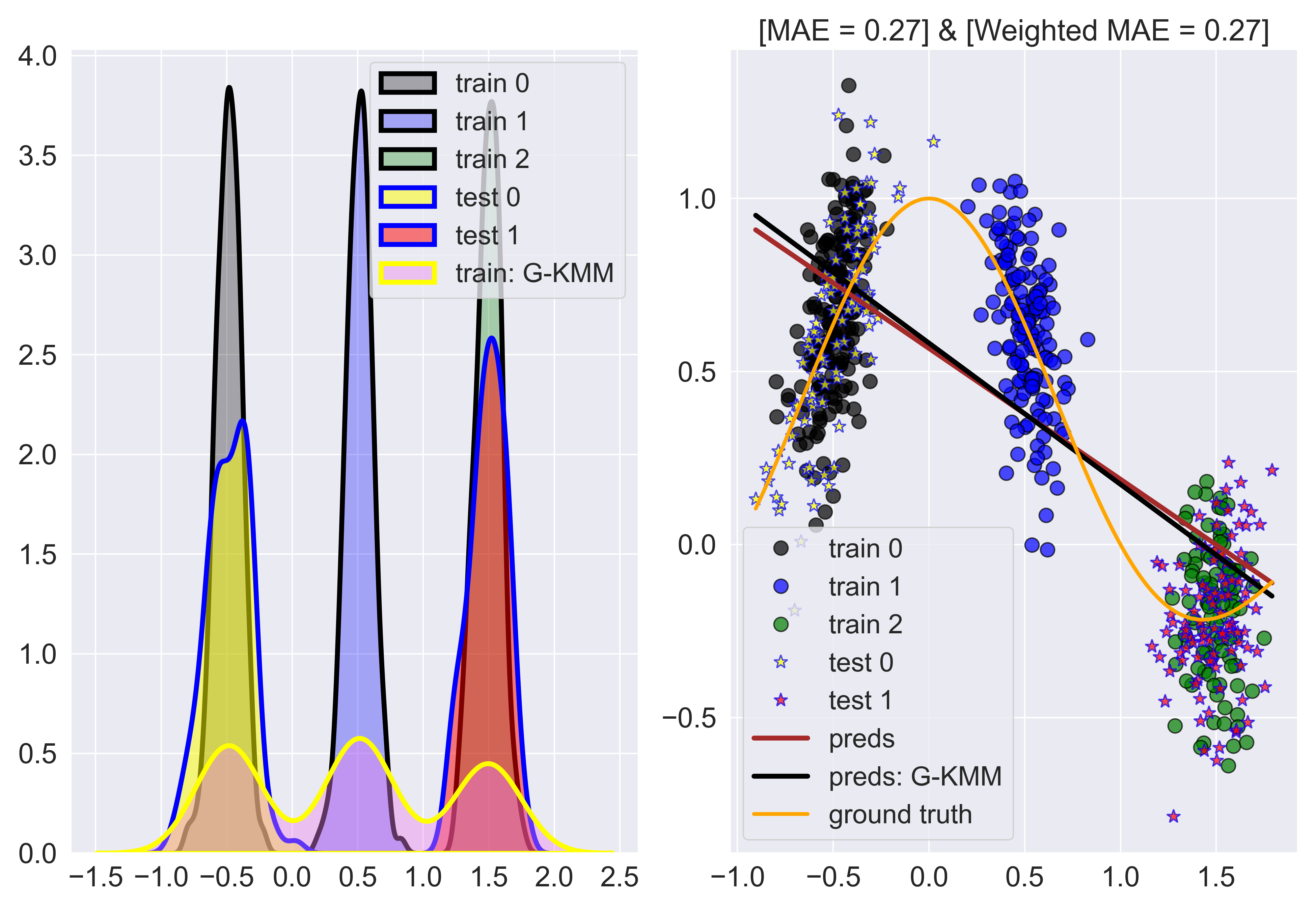}
    \end{subfigure}
    \begin{subfigure}[b]{0.44\textwidth}
        \includegraphics[scale=0.23]{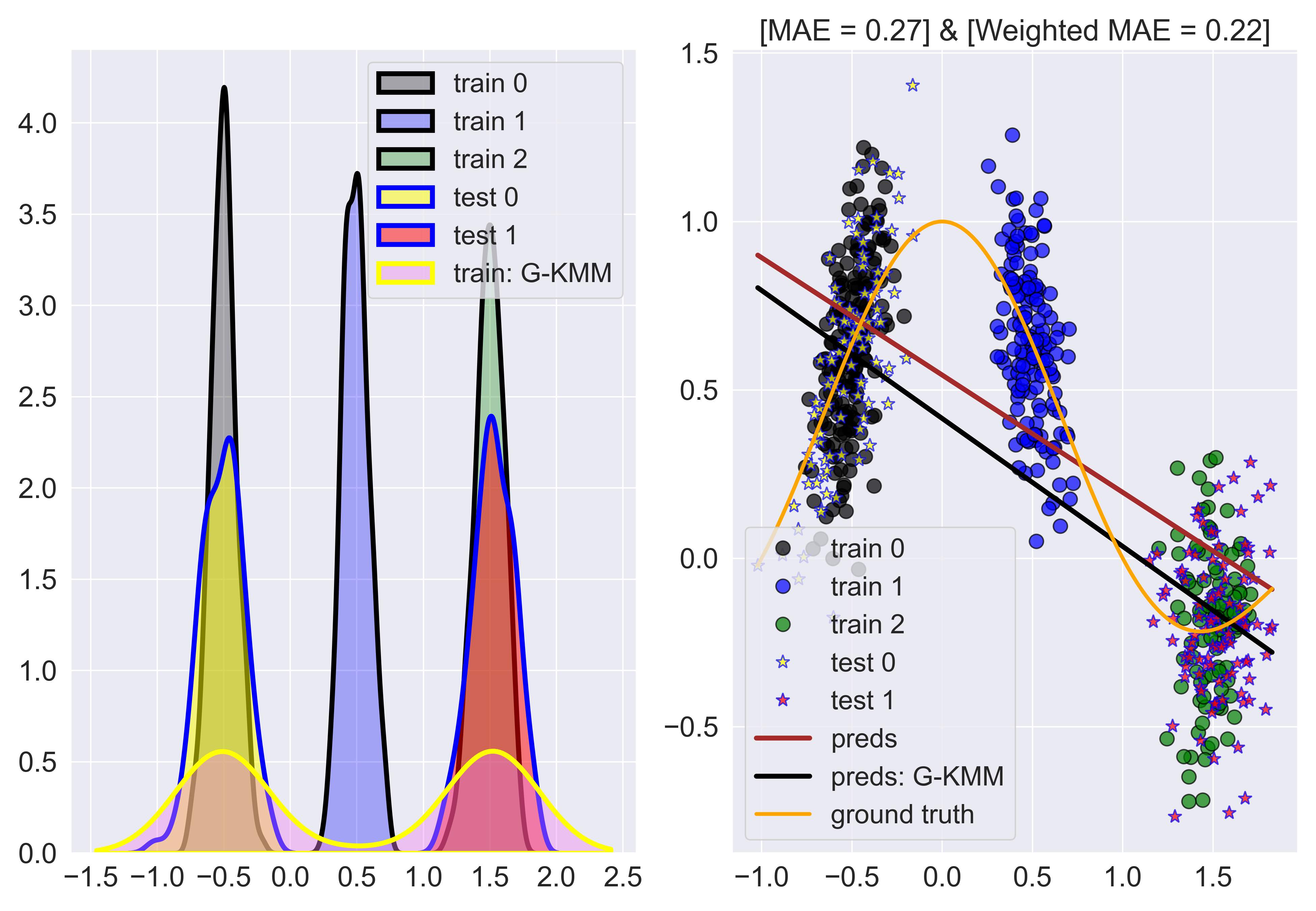}
    \end{subfigure}
    
    \begin{subfigure}[b]{0.44\textwidth}
        \includegraphics[scale=0.23]{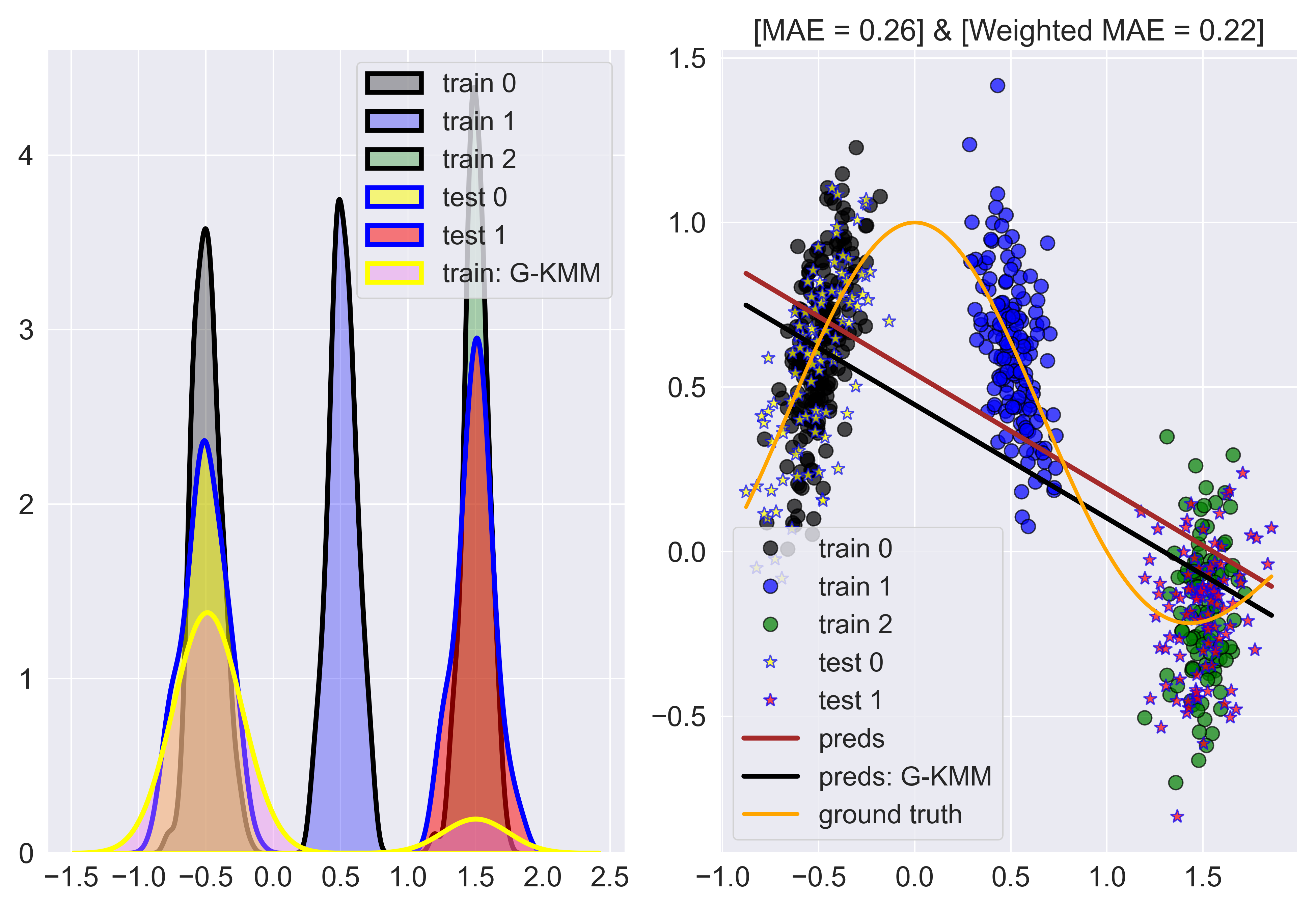}
    \end{subfigure}
    \begin{subfigure}[b]{0.44\textwidth}
        \includegraphics[scale=0.23]{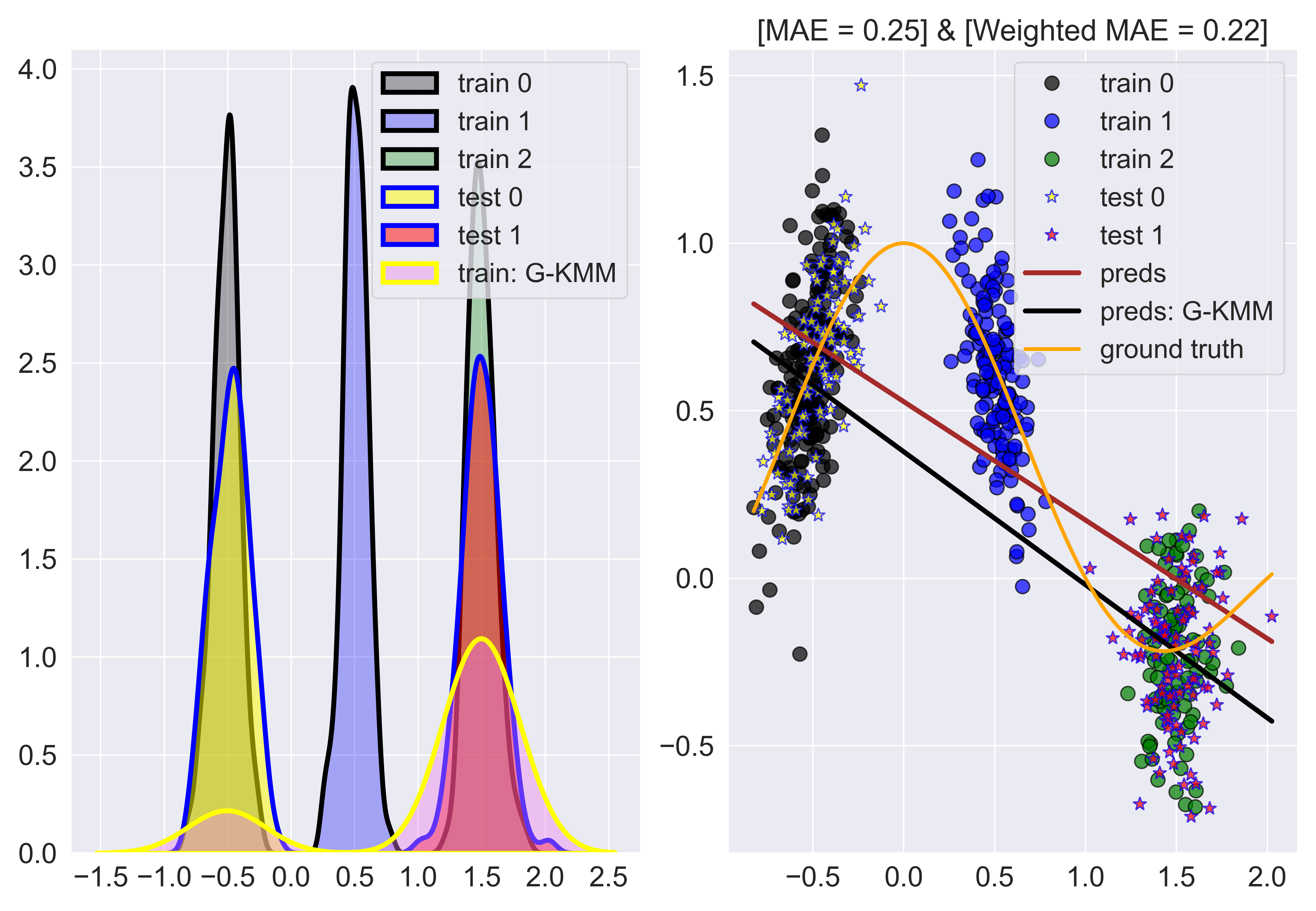}
    \end{subfigure}
    
\caption{Multiple train $\&$ test datasets}
\label{fig:4-GKMM-regression-train-and-test}
\end{figure}

Our next experiments are related to the comparison of various distributions by employing the cases of multiple train and test datasets. In figures \eqref{fig:2-GKMM-regression-train}, \eqref{fig:3-GKMM-regression-test} and \eqref{fig:4-GKMM-regression-train-and-test}, for each simulation that we have made, a custom selection of the train and test distributions is represented in the corresponding left plot while in the associated right plot we have considered visualizing the predictions of a basic \texttt{SGDRegressor} with and without the sample weights generated by the \texttt{Generalized KMM} algorithm. In order to inspect more closely the comparison of the effect of the density ratio weights, the title of each right plot shows the \textit{MAE} with and without the sample weights. For all the plots containing the regression results, the target is generated using a \textit{sinc} function at which we added a noise term following a normal distribution. Also, the $B$ term involved in \eqref{Empirical-Generalized-KMM} was set to $1000$. \\
The first simulation we have done is related to the case of multiple train datasets and it is shown in figure \eqref{fig:2-GKMM-regression-train}. For this, we have generated $3$ random normal train datasets with sizes $200$, $150$ and $100$, with the means $-0.5$, $0.5$ and $1.5$, and with the standard deviation equal to $0.1$. At the same time, the test dataset is composed of only $30$ samples generated using a normal distribution with mean $1.0$ and standard deviation $0.4$. In the left plots of the first row we have chosen a lower of value of $\sigma$, namely $0.1$ which implies that the weighted train distribution is uniformly distributed with respect to the train partitions. This can be easily visualized in the corresponding right plot where the weighted and unweighted predictions behave similarly. In the right pair of plots from the first row of figure \eqref{fig:2-GKMM-regression-train} we took $\sigma$ equal to $1.0$ but we have chosen the case of the $\alpha$-relative density ratio with $\alpha = 0.25$, and where the $\gamma_j$ weights of the training subsets were considered as $0.5$, $0.2$ and $0.05$, respectively. The effect of the $\alpha$-mixture density can be seen through the visualization of the distorsion of the weighted train distribution towards the skewed test dataset. For the last case, which is represented in the second row, we have set $\sigma$ to $100$, $\alpha$ to $0.5$ and the $\gamma_j$ values as $0.05$, $0.2$ and $0.25$, respectively. These choices shows a similar effect as in the previous case regarding the $\gamma_j$ mixture values.

Our next simulations presented in figure \eqref{fig:3-GKMM-regression-test} correspond to the case of multiple test datasets. We have generated a single train dataset of size $300$ from a normal distribution with mean $1.0$ and standard deviation $0.25$ for the results depicted in the first row, while for the second row the train was generated using a normal distribution with mean $0.5$ and standard deviation $0.25$, respectively. On the other hand, the test datasets, both of size $100$, were generated from normal distributions with means $-0.5$ and $1.5$, with the corresponding standard deviations equal to $0.15$. Furthermore, for all our simulations depicted in figure \eqref{fig:3-GKMM-regression-test} the parameter $\sigma$ was chosen as $100$. The left plots belonging to the first row shows that the weighted train distribution becomes closer to the test subset which overlaps the train dataset. On the other hand, the right plots from the first row shows the effect of the $\alpha$ mixture coefficient which was set to $0.75$ along with the $\gamma_1$ coefficient of the single train dataset which was eventually chosen as $0.25$. Here, we see that the mixture coefficient emphasize much more the test dataset which is closer to the training dataset, and it eventually leads to a worse approximation of the \texttt{SGDRegressor}. Finally, the simulation made in the plots from the second rows are based upon the same choice of the coefficents as in the previously described simulation, namely $\alpha$ is $0.75$, $\gamma_1$ was set to $0.25$ and $\sigma$ to $100$, respectively. The main difference is that the training dataset is shifted to the left hence it is located between the two test datasets. One can observe that the $\alpha$-mixture density ratio approach is suitable for this regression problem setting, by leading to a uniform-like distribution of the weighted train dataset.

The last experiment that we will present involves multiple train and also multiple test datasets and it is shown in figure \eqref{fig:4-GKMM-regression-train-and-test}. As before, we generate the train and test datasets using a random normal distribution. In the corresponding simulations we created $3$ train datasets of sizes $200$, $150$ and $100$, along with $2$ test datasets of sizes equal to $100$. The training subsets were generated from random normal distributions with means $-0.5$, $0.5$ and $1.5$, and standard deviation $0.1$. On the other hand, the two test datasets were generated using random normal distribution with means $-0.5$ and $1.5$, with a standard deviation equal to $0.15$. In the plots further to the left from the first row of \eqref{fig:4-GKMM-regression-train-and-test}, we have chosen the value $0.1$ for $\sigma$. Similar to the left-most plots from the first row of figure \eqref{fig:2-GKMM-regression-train}, the low value for $\sigma$ implies a weighted train distribution with $3$ peaks uniformly distributed, along with a weighted \textit{MAE} equal to the \textit{MAE} obtained from the unweighted predictions. On the other hand, in the right-most plots from the first row of \eqref{fig:4-GKMM-regression-train-and-test} the value for $\sigma$ was increased to $10$. This leads to only $2$ peaks, uniformly distributed and centered at the test distributions. Consequently, the \textit{MAE} metric decreases if one uses the weighted predictions of the \texttt{SGDRegressor}. \\
Now, let's turn our attention to the simulations made in the second row of figure \eqref{fig:4-GKMM-regression-train-and-test}. For the results shown in the further to the left plots we have chosen $\sigma$ equal to $100$, and the $\omega_i$ test weights $0.85$ and $0.15$, respectively. Along with these we have considered an $\alpha$-mixture density approach, where $\alpha$ was not defined directly, i.e. at first the $\gamma_j$ weights of the train subsets were constructed using the ratio of each train subset size and the size of the total training dataset, and then $\alpha$ was determined such that $\alpha$ and the sum of $\gamma_j$ add to the value of $1$ (for this see the basic example belonging to the ending part of section \eqref{sec:G-KMM_Motivation}). In this case one observes that the mixture density technique modifies the distribution of the peaks of the weighted training data. Furthermore, the value of the weights $\omega_i$ related to the test datasets shows that the higher the $\omega_i$ weight is then the higher is the peak pointing to the corresponding test dataset. Similar to the case of multiple test datasets which were depicted in the last row of figure \eqref{fig:3-GKMM-regression-test}, the $\alpha$-mixture density approach is crucial in the learning process of the optimal density ratio weights. \\
For the right-most plots shown in the second row of figure \eqref{fig:4-GKMM-regression-train-and-test} we considered also $\sigma$ equal to $100$. But, the weights corresponding to the training subsets, namely $\gamma_j$ were chosen this time as $0.25$, $0.2$ and $0.05$ while the weights $\omega_i$ for the test subsets were selected with the values $0.15$ and $0.85$, respectively. As explained before, since $\alpha$ and the sum of all the $\gamma_j$ coefficients must sum up to $1$, we set $\alpha$ to the value of $0.5$. Due to the fact that the weight of the second test subset is higher than the coefficient corresponding to the first test subset, the peak of the weighted train dataset is higher in the location of the second test subset. \\ \\
Finally, we conclude the present section by highlighting that the experiment made in figure \eqref{fig:1-GKMM-clusters} reveals a qualitative comparison between the classical KMM algorithm and our \texttt{Generalized KMM} density ratio optimization method. At the same time, the simulations presented in figures \eqref{fig:2-GKMM-regression-train}, \eqref{fig:3-GKMM-regression-test} and \eqref{fig:4-GKMM-regression-train-and-test} shows the versatility of our method through the choices of the coefficients, especially for the case of the $\alpha$-mixture density ratio.

\section{Conclusions $\&$ perspectives}\label{sec:G-KMM_Conclusions}
In this final section we present a brief overview of our \texttt{Generalized KMM} algorithm given through the \emph{quadratic optimization problem with constraints} \eqref{Empirical-Generalized-KMM} along with the underlying limitations and the possible extensions for future research. 

\subsection{Novelty}\label{subsec:G-KMM_Novelty}
In the present study, our main contribution is the introduction of a new type of density ratio estimation technique entitled \texttt{Generalized KMM}, which is an extension of the classical KMM algorithm. From both a theoretical and a practical point of view our proposed method is completely novel from the following perspectives:
\begin{itemize}
\item The \emph{Ensemble KMM} method from \cite{EnsembleKMM} is based on the idea of dividing the test dataset into multiple non-overlapping test sets, while \emph{Efficient Sampling KMM} introduced in \cite{EfficientSamplingKMM} is associated with the idea of a bootstrap aggregation approach for the training data. In contrast, our method is not developed using heuristic arguments, but it relies on the construction of a suitable loss function which attains its minimum value in the theoretical situation when one uses the true density ratio. 
\item In \cite{Gretton_KMM}, the classical KMM algorithm uses directly the density ratio model with respect to the training samples. But, in our work we employed the approach used in RuLSIF where the density ratio is approximated with a linear kernel model, where the underlying kernel depends on the test points. Hence we minimize a loss function with respect to some weights belonging to a lower-dimensional space, where the dimension is given by the total number of test samples.
\item Although we have constructed our minimization problem in connection to the cases consisting of non-overlapping train/test datasets, our approach contains as a particular case also a generalized version of the $\alpha$-relative density ratio, which is unique from the point of view of KMM-type methods. On the other hand, it is worth emphasizing that the theoretical construction of our method was done using the idea of non-overlapping sets, while the case of the $\alpha$-relative density ratio is devised only through a formal and mimetic approach.
\end{itemize}

\subsection{Research limitations}\label{subsec:G-KMM_Limitations}
Our method has not only advantages but it is also constrained by our inherent methodology as shown below:
\begin{itemize}
\item Despite the fact that the \texttt{Generalized KMM} is rigorously developed, one loses the parallelization property of the \emph{Ensemble KMM} and \emph{Efficient Sampling KMM}, respectively.
\item Similar to the classical KMM algorithm, our method has the same dependence on the hyper-parameters $\varepsilon$, $B$ and $\sigma$, respectively.
\end{itemize}

\subsection{Recommendations for future research}\label{subsec:G-KMM_Recommendations}
For future research, we propose the following methods to enlarge our KMM-type framework:
\begin{itemize}
\item In a similar manner with \cite{NeuralNetwork_KMM} we can extend our algorithm to the case of neural networks. More precisely, one can utilize the objective function given in \eqref{Empirical-Generalized-KMM} along with the constraints which can be applied directly into the forward propagation process. Consequently, we can make our method faster using randomized batch learning, hence we may utilize the KMM-type algorithm for adjusting the probability densities associated to data augmentation sample sets.
\item Similar to the classical KMM algorithm, our method is suitable for the estimation of the density ratio weights. In general, only a few training samples contribute to the reweighting process, due to the fact that the density ratio estimation and the regression/classification learning are separated. In order to alleviate this, we can proceed as in \cite{TailoringDRW} by simultaneously training our \texttt{Generalized KMM} density ratio model and the underlying weighted loss function, in the framework of supervised learning.
\end{itemize}

\bibliographystyle{plain}  
\bibliography{references}

\end{document}